\documentclass[lettersize,journal]{IEEEtran}
\usepackage{amsmath,amsfonts}
\usepackage{algorithmic}
\usepackage{algorithm}
\usepackage{array}
\usepackage{textcomp}
\usepackage{stfloats}
\usepackage{float}
\usepackage{url}
\usepackage{verbatim}
\usepackage{graphicx}
\usepackage{cite}
\usepackage{amssymb}
\usepackage{bm}
\usepackage[caption=false,font=normalsize,labelfont=sf,textfont=sf]{subfig}
\usepackage{booktabs}
\usepackage{bbding}
\usepackage{multirow}
\usepackage{xcolor}

\usepackage{booktabs}
\usepackage{multirow}
\usepackage{makecell}

\newcommand{\revred}[1]{#1}
\newcommand{\revblue}[1]{#1}
\newcommand{\tablenote}[1]{\par\vspace{3pt}\noindent\begin{minipage}{\linewidth}\footnotesize\raggedright\revblue{#1}\par\end{minipage}\par}

\usepackage[colorlinks=True,
            linkcolor=cyan,
            anchorcolor=blue,
            citecolor=blue,
            urlcolor=blue, 
            ]{hyperref}

\begin{document}

\title{MVSTF: Multi-View Spatio-Temporal Fusion for LiDAR-Based Moving Object Segmentation \\ 
}

\author{Jiaoju Zhou,
        Chao Ye\textsuperscript{*},
        Xuebo Yang,
        and Weiyang Lin
\thanks{
This work was supported in part by the National Natural Science Foundation of China under Grant 62503134 and 62373130. 
}
\thanks{
Jiaoju Zhou, Chao Ye, Xuebo Yang, and Weiyang Lin are with the Research Institute of Intelligent Control and Systems, Harbin Institute of Technology, Harbin 150001, China. (e-mail: zhoujiaoju@126.com, yechao@hit.edu.cn, wylin@hit.edu.cn, xueboyang@hit.edu.cn.)
}
\thanks{
\textsuperscript{*} Corresponding author(yechao@hit.edu.cn). 
}

}



\maketitle

\begin{abstract}
Autonomous mobile robots confront significant challenges in recognizing the motion properties of objects in dynamic environments.
However, existing methods are hindered by limitations such as boundary ambiguity, occlusion, viewpoint constraints, and training complexity.
In this paper, we propose an innovative end-to-end Multi-View Spatio-Temporal Fusion network, named \textbf{\textit{MVSTF}}, which utilizes sequential point clouds to generate multi-view images and feature differences, and designs a multi-branch network to extract spatio-temporal and dimensional features in parallel, and realizes high-precision moving object segmentation.
First, sequential point clouds are utilized to generate multi-view images and feature differences. 
Then, an end-to-end multi-branch network including Dual Down-sampling Blocks (DDB), Spatial Attention Blocks (SAB), Multi-Kernel Convolutional Neural Network blocks (MKCNN), Dual Feature Fusion Blocks (DFFB), and the Point Fusion (PF) module, is designed to extract motion features from feature differences and geometric and semantic features from multi-view images. 
Finally, an auxiliary semantic head is introduced to further improve the algorithm's robustness and the model's interpretability. 
This work is validated on the SemanticKITTI and KITTI-Road benchmarks, achieving state-of-the-art results on the SemanticKITTI-MOS validation set and the KITTI-Road validation set, and competitive results on the hidden SemanticKITTI-MOS test set. 
The code is available at \url{https://github.com/ZhouJJ111/MVSTF}.
\end{abstract}


\begin{figure}[!t]
    \centering
    \subfloat[LMNet]{
        \label{figure_a}
        \includegraphics[width = 0.96\linewidth,scale=1.0]{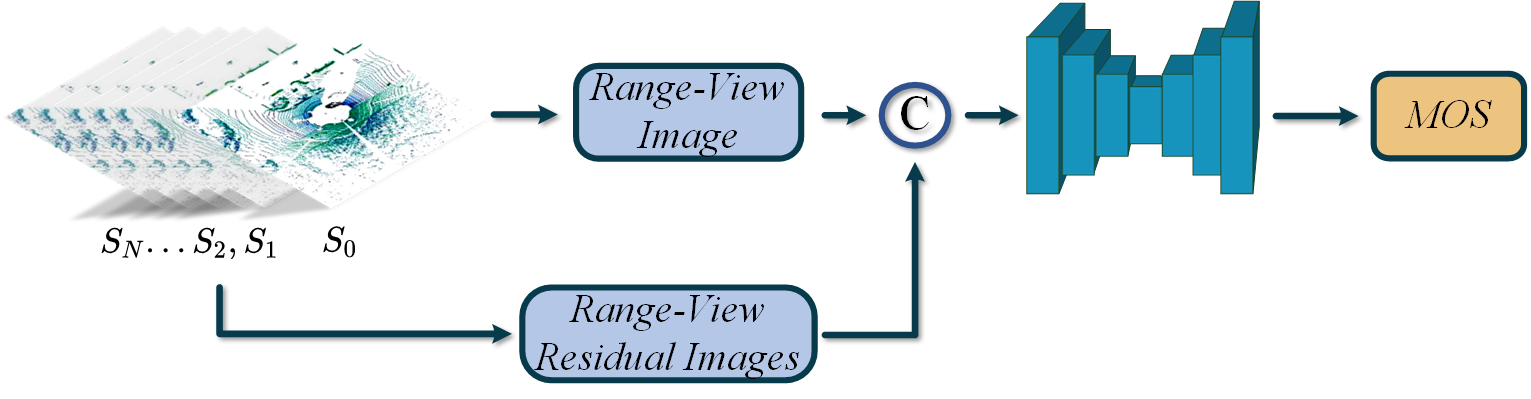}} \\
    \subfloat[MotionSeg3D]{
        \label{figure_b}
        \includegraphics[width = 0.96\linewidth,scale=1.0]{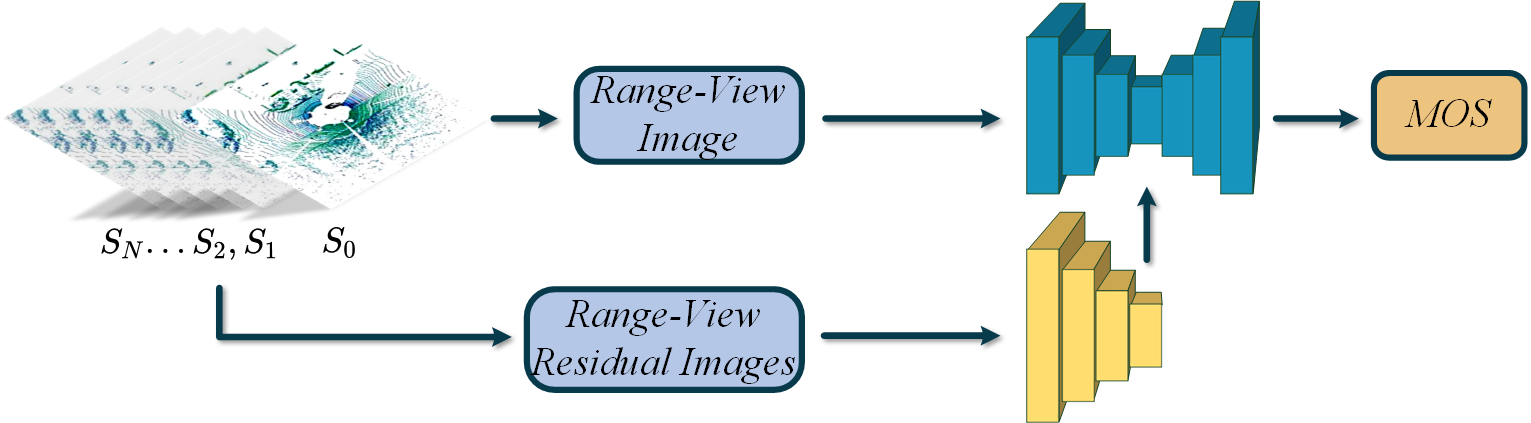}} \\
    \subfloat[MVSTF (ours)]{
        \label{figure_c}
         \includegraphics[width = 0.96\linewidth,scale=1.0]{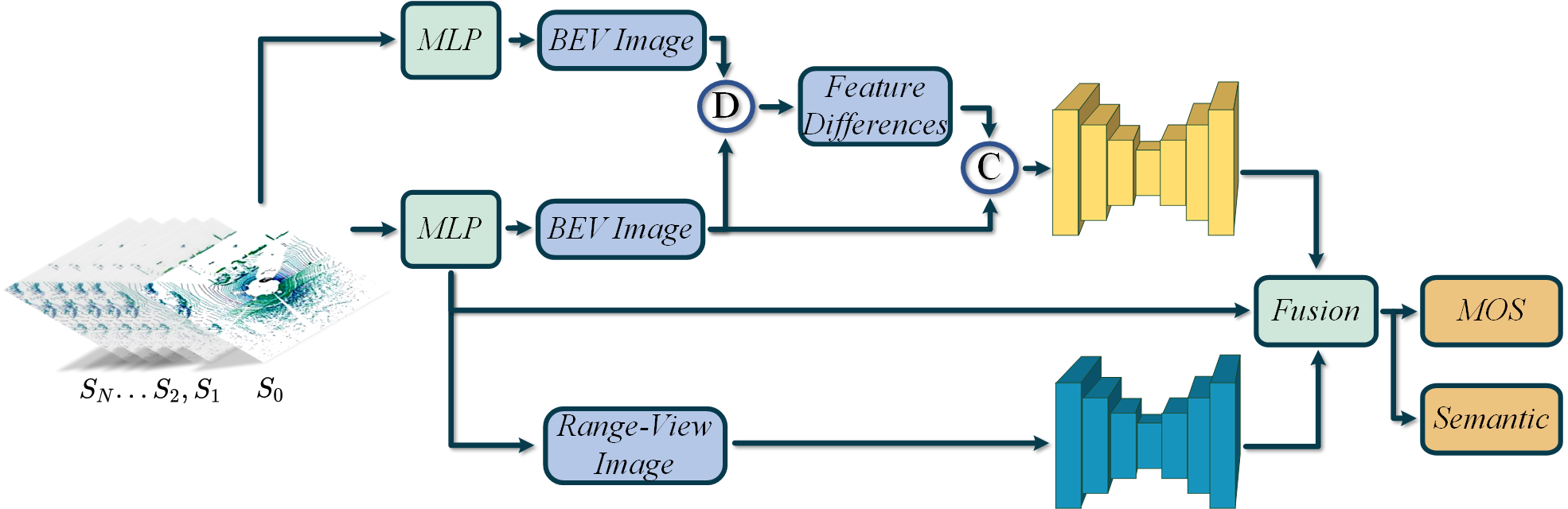}}
\caption{Comparative methods. (a) LMNet\cite{chen2021moving} integrates current range images from 3D LiDAR scans with prior residual images, enhancing feature extraction through temporal data integration. 
(b) MotionSeg3D\cite{sun2022efficient} employs a dual-branch architecture that separately processes range images and residuals to isolate appearance and motion features. 
(c) Unlike the single-view focus of LMNet\cite{chen2021moving} and MotionSeg3D\cite{sun2022efficient}, \textbf{\textit{MVSTF}} extracts multi-view spatio-temporal features from point clouds and utilizes an auxiliary semantic head to accelerate convergence to enhance network performance.}
\label{fig1}
\end{figure}

\section{Introduction}
\IEEEPARstart{U}{nderstanding} the dynamics of real-world environments is critical for safe and robust autonomous navigation.
As a core task in dynamic scene perception, Moving Object Segmentation (MOS) aims to distinguish dynamic agents (e.g., vehicles, pedestrians) from static background elements (e.g., buildings, parked cars), supporting modules such as Simultaneous Localization and Mapping (SLAM), path planning, and collision avoidance. 
In practice, autonomous systems often rely on the static-world assumption, expecting most elements to remain unchanged during observation. However, this assumption frequently breaks down in urban environments, where moving objects introduce ghost artifacts in maps, contaminate static representations, and can cause localization drift or even collisions. 
According to the U.S. National Highway Traffic Safety Administration (NHTSA)\cite{national2022traffic}, over $75\%$ of pedestrian fatalities in 2020 occurred outside intersections, often due to sudden road crossings that systems failed to detect in time. Recent work demonstrates that even microsecond-level motion cues can significantly improve emergency response \cite{wu2024moving}, highlighting the urgent need for temporally aware and spatially precise MOS frameworks. 
For instance, integrating an MOS module into the SuMa system reduces translation error by $20.5\%$ on the KITTI training set and up to $28.8\%$ in highly dynamic test regions \cite{chen2021moving}. 
Yet, the inherent sparsity, disorder, and non-uniformity of LiDAR point clouds pose significant challenges to reliable segmentation, especially under occlusions, viewpoint shifts, and complex dynamic scenes. 

Existing LiDAR-based MOS approaches can be broadly classified into four categories: point-based, voxel-based, projection-based, and hybrid methods. Point-based methods operate directly on the raw points and preserve fine-grained geometry, but they usually lack explicit temporal modeling. Voxel-based methods (e.g., 4DMOS\cite{mersch2022receding}, InMOS\cite{wang2023insmos}) discretize the point cloud into sparse 4D voxels and employ sparse 4D convolutional networks to extract spatiotemporal features efficiently. These methods offer strong capabilities in modeling object trajectories and capturing local geometric structures. However, the voxelization process introduces quantization artifacts that degrade fine-grained detail representation, and their computational and memory demands limit scalability in large-scale dynamic scenes. 
In contrast, projection-based methods transform the 3D point cloud into 2D images to leverage mature image segmentation architectures. Range View (RV)-based approaches (e.g., LMNet\cite{chen2021moving}, MotionSeg3D\cite{sun2022efficient}, RVMOS\cite{kim2022rvmos}) project the point cloud into depth maps and compute residuals between consecutive frames to infer motion cues. Bird's-Eye View (BEV)-based methods (e.g., MotionBEV\cite{zhou2023motionbev}) encode planar object motion from a top-down perspective.
Despite their computational efficiency, these methods suffer from occlusion, geometric distortion, and boundary ambiguity due to reliance on a single viewpoint. Moreover, they often require heuristic post-processing steps (e.g., K-Nearest Neighbors (KNN) or Conditional Random Fields (CRF)) to project 2D predictions back to the 3D domain, introducing additional complexity and uncertainty into the pipeline. Hybrid methods combine multiple representations, such as point, voxel, or projection features, to exploit their complementary strengths at the cost of higher architectural complexity. A detailed survey of all four categories is provided in Section II.

\begin{figure}[!t]
\centering    
\includegraphics[width = 0.96\linewidth,scale=1.0]{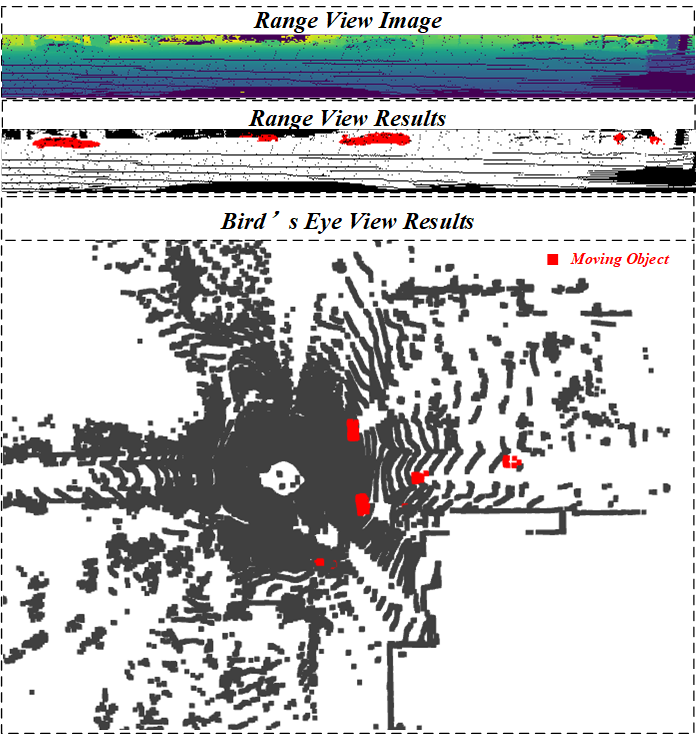}
\caption{The Multi-view representations: range view and bird's-eye view.}
\label{fig2}
\end{figure}

To address the aforementioned limitations, we propose MVSTF, a unified framework that integrates multi-view representations and temporal modeling to enhance the robustness and accuracy of MOS. The design of MVSTF is guided by five core principles: 
(1) Multi-view complementarity: By combining spherical and bird's-eye projections, MVSTF mitigates information loss and occlusion caused by single-view limitations, enabling comprehensive geometric perception; 
(2) Multi-representation fusion: A multi-path architecture, composed of a point-based MLP branch and a multi-view CNN branch, is designed to balance fine-grained geometry with global semantic context; 
(3) Explicit motion modeling: Inter-frame feature differences are computed on BEV representations to capture structured and robust motion cues along dominant object trajectories; 
(4) End-to-end optimization: A Point Fusion (PF) module eliminates reliance on post-processing steps, enabling a fully differentiable training pipeline with joint optimization across stages; 
(5) Auxiliary semantic supervision: An auxiliary semantic head is incorporated during training to inject static structural priors into the motion prediction process, significantly improving generalization in dynamic and cluttered scenes. 
Together, these components form a cohesive and high-performance framework tailored to real-world complexity. 
As illustrated in Fig.~\ref{fig1}, MVSTF integrates four complementary representations---raw point cloud, RV, BEV, and inter-frame feature differences---into a unified perception pipeline. 
Comparative and ablation studies on the SemanticKITTI and KITTI-Road benchmarks demonstrate the effectiveness of this approach. 
The multi-view representations and segmentation results of point clouds are shown in Fig.~\ref{fig2}. 
In summary, the main contributions of this paper are:
\begin{itemize}
\item We propose a novel end-to-end framework, MVSTF, that leverages multi-view representations and feature differences in point cloud sequences for spatio-temporal fusion. This approach effectively addresses boundary-blurring, occlusion, and training complexity, significantly improving the performance of the MOS task. 
\item The projection branches incorporate Dual Down-sampling Blocks (DDB), Spatial Attention Blocks (SAB), Multi-Kernel Convolutional Neural Network blocks (MKCNN), and Dual Feature Fusion Blocks (DFFB) to extract multi-scale spatio-temporal features. The MKCNN block, illustrated in Fig.~\ref{fig4}, employs directionally decomposed convolutions with kernels of different sizes to capture multi-scale motion and appearance cues, particularly for distant and small objects. 
The Point Fusion (PF) module then integrates multi-view and point-based features, preserving fine-grained geometric, semantic, and spatio-temporal information. 
\item The introduction of the auxiliary semantic head significantly enhances the network's learning capacity and model robustness, without increasing extra inference parameters. 
\item \textbf{\textit{MVSTF}} achieves state-of-the-art performance on the SemanticKITTI-MOS validation set and the KITTI-Road validation set, and remains competitive on the hidden SemanticKITTI-MOS test set under the combined training protocol. 
\end{itemize}

\section{Related Work}
Moving Object Segmentation (MOS) has become a key task in autonomous driving and robotics, aiming to capture both spatial geometry and temporal motion of dynamic agents. Unlike semantic segmentation, which focuses on static class labels, MOS emphasizes the temporal evolution of point states. 
To address the increasing demand for spatiotemporal modeling, LiDAR-based MOS methods have evolved from semantic segmentation architectures by adopting diverse input representations. 
Based on the type of input representation, existing approaches can be broadly categorized into four groups: point-based, voxel-based, projection-based, and hybrid methods. Each category reflects a distinct trade-off between spatial representation fidelity and temporal modeling capacity. 

\subsection{Point-based methods}
Point-based methods operate directly on raw 3D point clouds, preserving fine-grained geometric structures without the quantization loss introduced by voxelization or projection. 
Foundational approaches such as PointNet\cite{qi2017pointnet} and PointNet++\cite{qi2017pointnet++} leverage shared multilayer perceptrons (MLPs) and hierarchical feature aggregation to enable effective point-level processing. 
Subsequent methods like KPConv\cite{thomas2019kpconv} and PointConv\cite{wu2019pointconv} introduce irregular convolutional kernels tailored to the unordered nature of point clouds, while graph-based models such as GACNet\cite{wang2019graph} enhance local contextual reasoning through attention mechanisms. 
For improved efficiency, lightweight models like RandLA-Net\cite{hu2020randla} utilize random sampling and sparse neighborhood construction to reduce computational cost. 

Despite their success in static point cloud segmentation, point-based methods face two key limitations in dynamic environments. First, their high computational and memory costs, combined with inconsistent neighborhood definitions, hinder their scalability in large-scale or real-time applications. Second---and more critically---most of these methods are not equipped with explicit temporal modeling mechanisms, making them inadequate for capturing inter-frame motion cues, which are essential for accurate and robust MOS. 

\subsection{Voxel-based methods}

Voxel-based methods discretize irregular point clouds into structured 3D voxel grids, enabling convolutional neural networks to efficiently capture spatial structures and semantic context. In static semantic segmentation, MinkowskiNet\cite{choy20194d} introduces a sparse tensor framework and 4D sparse convolutions to model large-scale scenes with high computational efficiency. SPVCNN\cite{tang2020searching} further refines voxel-based operations to reduce overhead while maintaining accuracy, and Cylinder3D\cite{zhu2021cylindrical} employs cylindrical voxelization with asymmetric convolutions to adapt to the non-uniform density patterns of LiDAR data. 
In the context of MOS, voxel-based architectures have been extended to capture spatiotemporal dynamics. 
For instance, SpSequenceNet\cite{shi2020spsequencenet} voxelizes sequential scans and introduces cross-frame global attention (CGA) and cross-frame local interpolation (CLI) to extract temporal dependencies. 
4DMOS\cite{mersch2022receding} constructs sparse 4D tensors and applies Bayesian filtering for robust temporal fusion. 
InMOS\cite{wang2023insmos} enhances instance-level segmentation by integrating an instance-awareness module into the voxel backbone.

Despite their structural advantages, voxel-based methods face two key limitations. First, voxelization inevitably introduces quantization errors, degrading the representation of fine-grained geometries---especially in sparse or cluttered regions. Second, voxel granularity poses a critical trade-off: finer resolutions improve spatial fidelity but increase computational cost, while coarser resolutions enhance efficiency at the expense of detail. Balancing accuracy, efficiency, and scalability remains an open challenge for deploying voxel-based MOS in real-world dynamic environments.

\subsection{Projection-based methods}

Projection-based methods transform irregular 3D point clouds into structured 2D representations---such as RV, BEV, or cylindrical projections---enabling efficient feature extraction via convolutional neural networks. 
These methods are widely adopted in LiDAR semantic segmentation due to their regular data structures and computational efficiency. 
For instance, LeNet\cite{ding2023lenet} \revred{introduces boundary-aware} filtering, sparse convolutions, and interpolation strategies to enhance segmentation accuracy and runtime. 
In the MOS domain, projection-based methods are extended to capture temporal dynamics. 
LMNet\cite{chen2021moving} encodes inter-frame residuals between current and historical RV images to infer motion. 
MotionSeg3D\cite{sun2022efficient} adopts a dual-branch spatiotemporal architecture with motion-guided attention and point-wise refinement to reduce ghosting. 
RVMOS\cite{kim2022rvmos} integrates temporal interval modulation and augmentation strategies to improve robustness. 
MotionBEV\cite{zhou2023motionbev} leverages a polar BEV projection and height-difference-based motion encoding, augmented with cross-attentional fusion between appearance and motion features for enhanced dynamic segmentation. 

Despite their efficiency, projection-based methods suffer from inherent limitations due to their reliance on single viewpoints. RV projections are prone to spherical distortion, which affects near-field object geometry, while BEV projections struggle with uneven vertical point density and occlusions, leading to missing spatial context. These issues are especially critical for MOS, where precise boundary localization and temporal consistency are essential. 
To address these limitations, we propose MVSTF, a multi-view spatiotemporal fusion framework that integrates complementary spherical and BEV projections. 
By combining geometric cues from different perspectives and explicitly modeling inter-frame feature differences, MVSTF constructs a unified, modular, and temporally-aware framework for robust motion segmentation in complex real-world environments.

\subsection{Hybrid methods}
Hybrid approaches aim to integrate multiple point cloud representations---including point-wise, voxel-based, and projection-based features---to leverage their complementary geometric and semantic cues for robust spatial-temporal modeling in dynamic environments.
In static semantic segmentation, representative methods such as RPVNet\cite{xu2021rpvnet} introduce a gated fusion mechanism to combine RV, point, and voxel features for efficient multi-representation learning. PRNet\cite{li2022prnet} further achieves a favorable trade-off between modeling accuracy and computational efficiency by fusing point and RV features under a lightweight architecture. 
For MOS, recent works have increasingly explored the joint modeling of multi-representations and temporal dynamics. 
SMVF\cite{zhang2022sequential} adopts a multi-branch architecture that integrates raw point clouds with multi-view projection features, and introduces a Sequential Instance Copy-Paste (SICP) module to enhance dynamic sample representation. 
SegNet4D\cite{wang2024segnet4d} decouples semantic and motion branches, fusing point and BEV features to support multi-task learning. 
Mars3D\cite{liu2023mars3d} explicitly models motion trajectories via inter-frame residual features, significantly improving the temporal consistency and recognition of dynamic objects. 
Recent studies have advanced LiDAR scene understanding through temporal modeling\cite{wu2024taseg}, weak supervision\cite{chen2025weakly}, and semi-supervised\cite{wu2024semi} learning. Point cloud perception has also been extended to object-level SLAM\cite{li2025dqo} and category-level robotic grasping\cite{ye2025csgrasp}. While these approaches improve generalization and reduce annotation costs, they often rely on heuristic designs or limited supervision. 
In contrast, MVSTF adopts a fully supervised, end-to-end framework that achieves a balanced trade-off among accuracy, computational efficiency, and deployability, making it well-suited for motion segmentation in complex dynamic environments. 

Despite their effectiveness, hybrid methods often suffer from architectural complexity, high computational cost,  and opaque fusion strategies, which hinder real-world deployment.
To address these issues, MVSTF introduces a multi-branch framework that fuses point-based and projection-based features, 
with the following three key innovations: 
(1) Architectural refinement: Inspired by SegNet4D\cite{wang2024segnet4d}, MVSTF adopts a dual-branch structure that jointly learns semantic categories and motion states, improving discriminability and generalization across diverse environments. 
(2) Explicit temporal modeling: Building on insights from Mars3D\cite{liu2023mars3d}, MVSTF introduces inter-frame feature differencing in the BEV branch to capture motion cues along the ground-dominant plane, improving temporal consistency and motion discrimination. 
(3) Dual temporal strategies: MVSTF supports two temporal modeling modes to accommodate different application needs: (i) a feature difference-based strategy for high-precision offline tasks such as HD map generation and trajectory prediction, and (ii) a height residual-based mode that enables real-time inference at sensor frame rates, suitable for vehicle-mounted or edge-computing platforms. 
\section{Our approach}
\begin{figure*}[!t]
  \centering    
  \includegraphics[width = \linewidth,scale=0.8000]{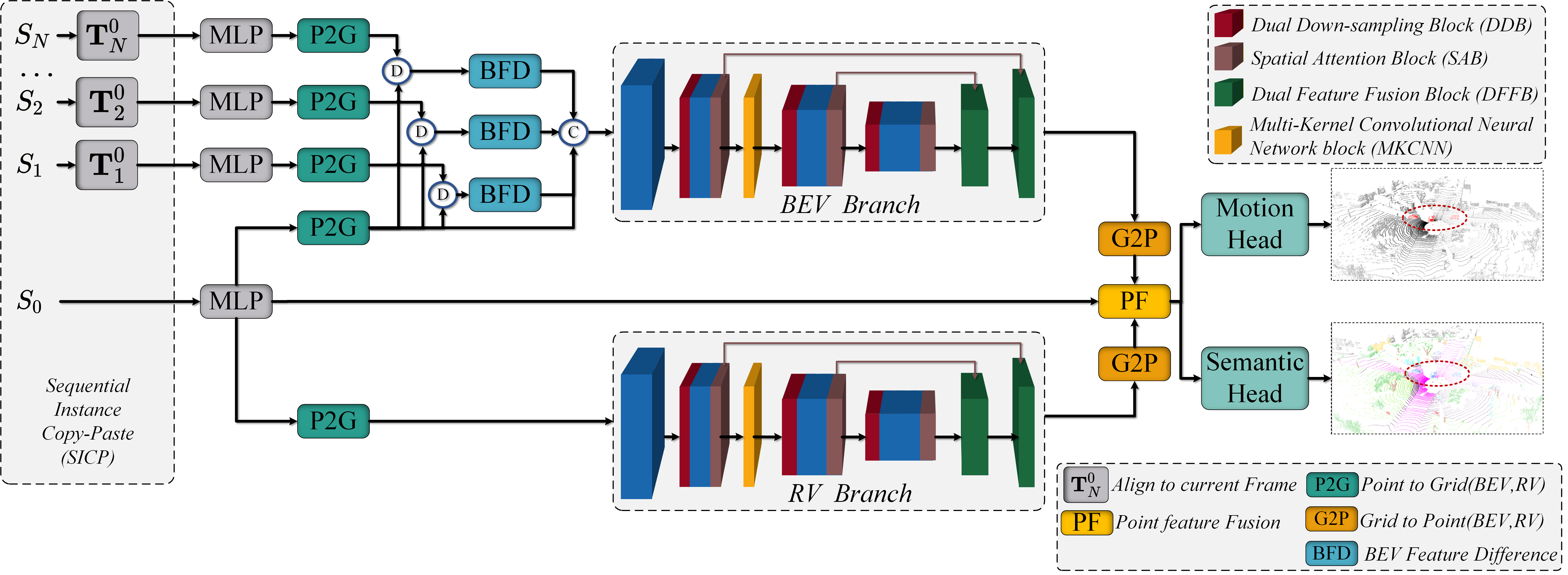}
\caption{Overview of \textbf{\textit{MVSTF}}. 
Multi-view feature encoding: 
We convert the point cloud into BEV and RV images, perform ego-motion compensation on the historical frames, and calculate BEV feature differences. 
Multi-Branch Networks: The Fully Convolutional Network (FCN) consists of Dual Down-sampling Blocks (DDB), Spatial Attention Blocks (SAB), Multi-Kernel Convolutional Neural Network blocks (MKCNN), Dual Feature Fusion Blocks (DFFB), and Point Fusion (PF) modules to enable end-to-end network training.
In the training phase, the auxiliary semantic head promotes parameter convergence and remarkably improves network performance, and we remove the auxiliary semantic head in the inference phase.
The BEV feature difference module takes the current and historical BEV feature maps as input, computes their differences, and outputs the concatenation of the current-frame features and the feature differences, which is then fed into the BEV branch.}
\label{fig3}
\end{figure*}
As shown in Fig.~\ref{fig3}, MVSTF is organized as three parallel feature branches that encode different representations of the same point cloud sequence: a point-based branch processes the raw points with a shared MLP, a BEV branch takes as input the current BEV image concatenated with the inter-frame feature differences, and an RV branch takes as input the current range image. In each projection branch, the encoder is composed of three building blocks with distinct roles: the Dual Down-sampling Blocks (DDB) progressively reduce the spatial resolution while retaining fine-grained details, the Spatial Attention Blocks (SAB) emphasize informative locations at each scale, and the Multi-Kernel Convolutional Neural Network blocks (MKCNN) capture multi-scale motion and appearance patterns. In the decoder, the Dual Feature Fusion Blocks (DFFB) adaptively fuse features from different layers. Finally, the Point Fusion (PF) module maps the BEV and RV features back to the original points and merges them with the point-based features to produce the final point-wise prediction. During training, an auxiliary semantic head provides additional supervision and is removed at inference. Fig.~\ref{fig1}(c) presents the concept-level pipeline of MVSTF, while Fig.~\ref{fig3} provides the module-level architecture; the multi-view images, the feature differences, and the auxiliary semantic head in Fig.~\ref{fig1}(c) correspond, respectively, to the projection branches, the BEV feature difference module, and the auxiliary semantic head in Fig.~\ref{fig3}.

The BEV and RV representations of point clouds could provide complementary information, effectively addressing the spatial ambiguities inherent in single-view approaches. Thus, we employ a multi-view strategy to separately extract motion and spatial features. 
To mitigate information loss during the projection process and address boundary-blurring from back-projection, we introduce a feature extraction branch for raw point clouds. This enhancement significantly improves segmentation accuracy, particularly for small objects. 
\subsection{Multi-View Feature Encoding}
The data preprocessing module processes raw point clouds using the PointNet \cite{qi2017pointnet} network to extract point-level features. It then combines the BEV feature differences from previous scans with the current scan's BEV image, feeding them into the BEV branch. Simultaneously, the RV image from the current scan is input to the RV branch. 
The three-branch network operates independently. 
To mitigate the challenges posed by limited sample size and occlusion of moving objects, we adopt the SICP strategy for data augmentation. 

Given a 3D point cloud sequence $ S= \{S_{j}\}_{j=0}^{N}$ of $N+1$ LiDAR scans, with the current scan denoted as $S_{0}$. 
Assume that the scan $S_{j}$ contains $M_{j}$ points, where the homogeneous coordinate of the $i$-th point is $ \mathbf{p}_{i} = [x_{i},y_{i},z_{i},1]^{T}$. 
Our goal is to predict the dynamic properties of point clouds in the current scan, i.e., whether it is moving or not.

\subsubsection{Ego-Motion Compensation}
Capturing temporal and motion information from sequential LiDAR scans is crucial for accurate analysis. Since each scan is defined in its local coordinate system, relative pose transformation matrices from SLAM or odometry are required to align historical scans with the current coordinate system, enabling spherical and BEV projections. 
\begin{equation}
  \begin{aligned}
  & S_{j}^{0} = \{\mathbf{T}_{j}^{0} \mathbf{p}_{i} | \mathbf{p}_{i} \in S_{j}, j = 1,\cdots, N \}, \\
  & \text{with } \mathbf{T}_{j}^{0} = \prod_{k=0}^{j-1} \mathbf{T}_{j-k}^{j-k-1},
  \end{aligned}
\end{equation}
where $ \mathbf{T}_{j-k}^{j-k-1} \in \mathbb{R}^{4*4}$ indicates the relative transformation between scans.


\subsubsection{Point to Grid}
To achieve multi-view spatial structure encoding, we employ a point-to-grid projection operation to map the original 3D point cloud onto the spherical and BEV planes, respectively, to generate structured 2D features

\textbf{Spherical Projection.} 
For the spherical projection, the point cloud is transformed into a RV image according to the following formulation:
\begin{equation}
  \begin{pmatrix} u_{i} \\ v_{i} \end{pmatrix} =\begin{pmatrix} \frac{1}{2}  *\left[\left(1-\arctan (y_{i}, x_{i}) / \pi\right)\right] \cdot W_{rv} \\\left[1-\left(\arcsin \left(z_{i}/r_{i} \right) + f_{\text {up}}\right) / f\right] \cdot H_{rv}\end{pmatrix},
\end{equation}
where $W_{rv}$ and $H_{rv}$ denote the width and height of the RV image, respectively. $r_{i}=\|\mathbf{P}_{i}\|_2=\sqrt{x_{i}^2+y_{i}^2+z_{i}^2}$ denotes the range value of each point, and $ f = f_{\text {up}}+ f_{\text {down}} $ denotes the vertical field-of-view of the LiDAR sensor. 
Each coordinate $(u_{i},v_{i})$ can index a 5-dimensional feature of the corresponding point, containing 3D coordinates $(x_{i},y_{i},z_{i})$, range $r_{i}$, and intensity $e_{i}$. Since the projection is a many-to-one mapping, multiple points may fall into the same pixel; in this case, we aggregate their features by element-wise max pooling within the pixel, so that each pixel of the RV image contains a single feature vector.

\textbf{BEV projection.}
For the BEV projection, the point cloud is mapped onto the x-y plane to construct a BEV representation as follows: 
\begin{equation}
  \begin{pmatrix}u_{i} \\v_{i}\end{pmatrix} = \begin{pmatrix}\frac{x_{i}-x_{min}}{x_{max}-x_{min}} \cdot W_{bev} \\\frac{y_{i}-y_{min}}{y_{max}-y_{min}} \cdot H_{bev}\end{pmatrix},
\end{equation}
where $W_{bev}$ and $H_{bev}$ denote the width and height of the BEV image, and $(x_{min}, y_{min}, x_{max}, y_{max})$ represents the discrete range of a 2D X-Y planar rectangular grid.


\subsubsection{Motion Features Generation}
To extract motion information from successive LiDAR scans, RV images are sensitive to distance variations, and most objects primarily move within the X-Y plane. Hence, we propose extracting motion features from BEV feature differences, which more effectively capture dynamic object changes.

\textbf{Height Residuals.} 
Similar to the RV residuals in LMNet\cite{chen2021moving}, the BEV height residual image is derived in three steps: First, the previous scans are transformed into the current local coordinate system using ego-motion compensation. Next, the past scans are projected onto the current BEV image. Finally, the residuals for each pixel in the BEV image are computed by calculating the height difference between the current frame's BEV image and the transformed frame. 
\begin{equation}
    d_{j,i}^{0} = \begin{cases} \left \| z_{i}^{0}  - z_{j,i}^{0} \right \|, &  i \in \text{valid pixels} \\ 0 & \text{ otherwise}\end{cases},
\end{equation}
where $z_{i}^{0}$ denotes the height value of the current frame in BEV coordinates $(u_{i},v_{i})$ by max pooling and 
$z_{j,i}^{0}$ represents the height value corresponding to the transformation from the $j$-th frame to the current frame in the same BEV coordinates. 
The height residual images of the historical frames and the current frame are then concatenated along the channel dimension. 
Each pixel of the fused BEV image contains the vector information as $(x_{i},y_{i},z_{i},r_{i},e_{i},\delta x_{i},\delta y_{i},d_{1,i}^{0},\cdots,d_{j,i}^{0}\cdots,d_{N,i}^{0})$, 
where $(x_{i},y_{i},z_{i})$ and $e_{i}$ denote the coordinates and intensity values of point $\mathbf{p}_{i}$, respectively.
$r_{i}$ represents the range value, and $N+1$ denotes the size of the time window. 
$(\delta x_{i},\delta y_{i})$ denotes the difference from the point to the current grid, 
and $d_{j}^{0}$ indicates the height residual of the BEV image between the last ${j}$-th frame and the current frame.

\textbf{Feature Differences.} 
First, we select $N$ historical frames and the current frame as inputs. The same MLP with shared weights is applied independently to each scan to extract point-level features, and the resulting features are subsequently transformed into the BEV space by the point-to-grid operation. 
\begin{equation}
\{ \mathcal{U}_i \}_{i=0}^N = \left\{ P2G(MLP(\mathcal{S}_i^{0})) \right\}_{i=0}^N.
\end{equation}
After that, the feature differences between the historical frames and the current frame in BEV space are calculated, aiming to offset static objects while highlighting the feature magnitude of moving objects. 
\begin{equation}
\{ \mathcal{D}_i \}_{i=1}^{N} = \{ \mathcal{U}_i - \mathcal{U}_0 \}_{i=1}^{N}.
\end{equation}
Finally, the current frame features $\mathcal{U}_0$ and the feature differences from the $N$ historical frames $\{\mathcal{D}_i \}_{i=1}^{N}$ are concatenated along the channel dimension. 
Because the subtraction is performed after the learned MLP embedding and BEV projection, the resulting difference is a learned feature residual rather than a direct geometric residual. Nevertheless, after ego-motion compensation, static content is largely canceled by the subtraction, while motion-induced changes persist in feature space. The residual therefore provides an effective motion-oriented cue for the BEV branch.
\begin{figure}[!t]
    \centering    
    \includegraphics[width = 0.85\linewidth,scale=1.0]{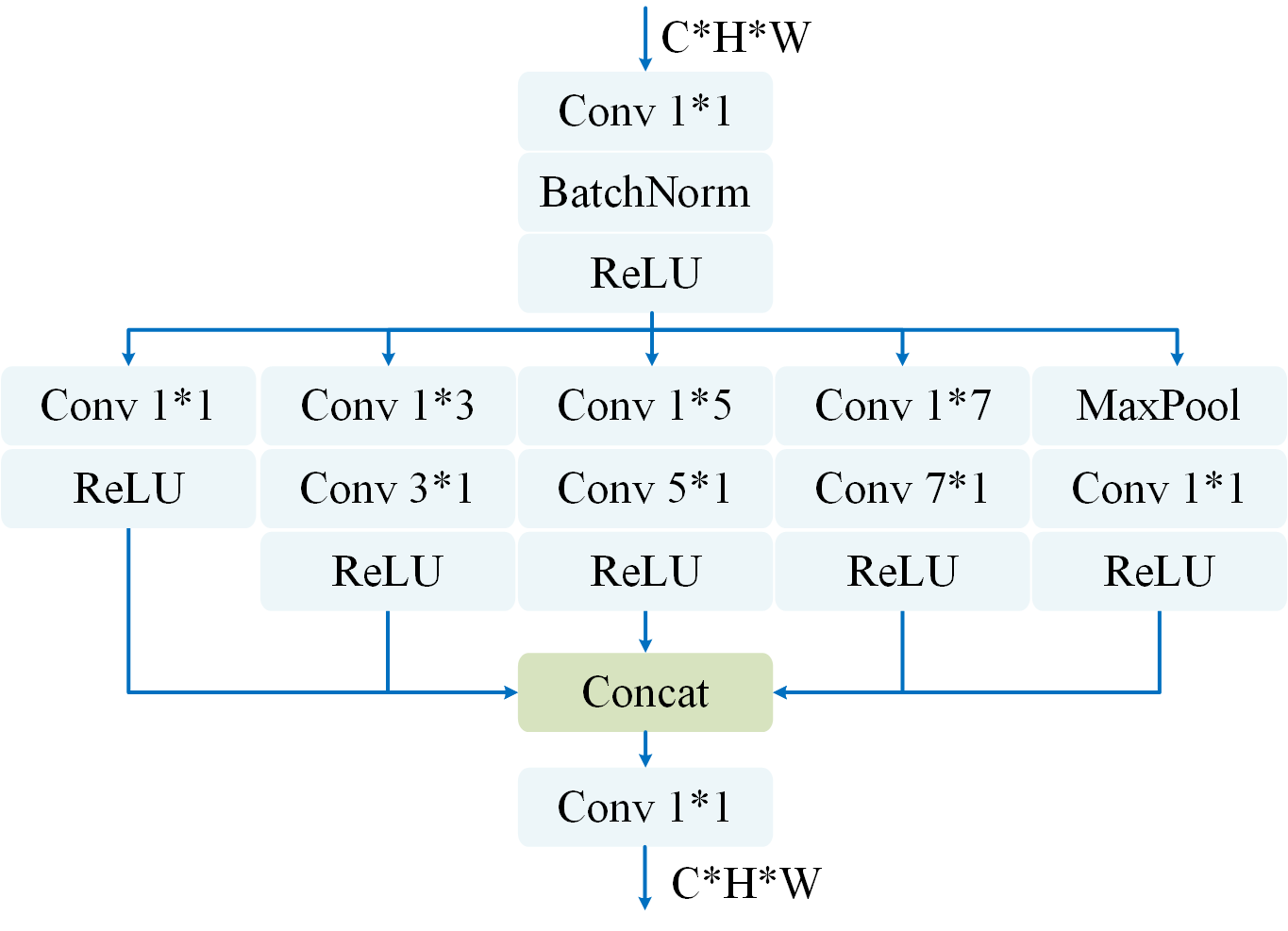}
\caption{The Multi-Kernel Convolutional Neural Network block (MKCNN).}
\label{fig4}
\end{figure}

\begin{figure}[!t]
    \centering
    \subfloat[The Spatial Attention Block]{
        \label{figure_5a}
        \includegraphics[width = 0.85\linewidth,scale=0.3]{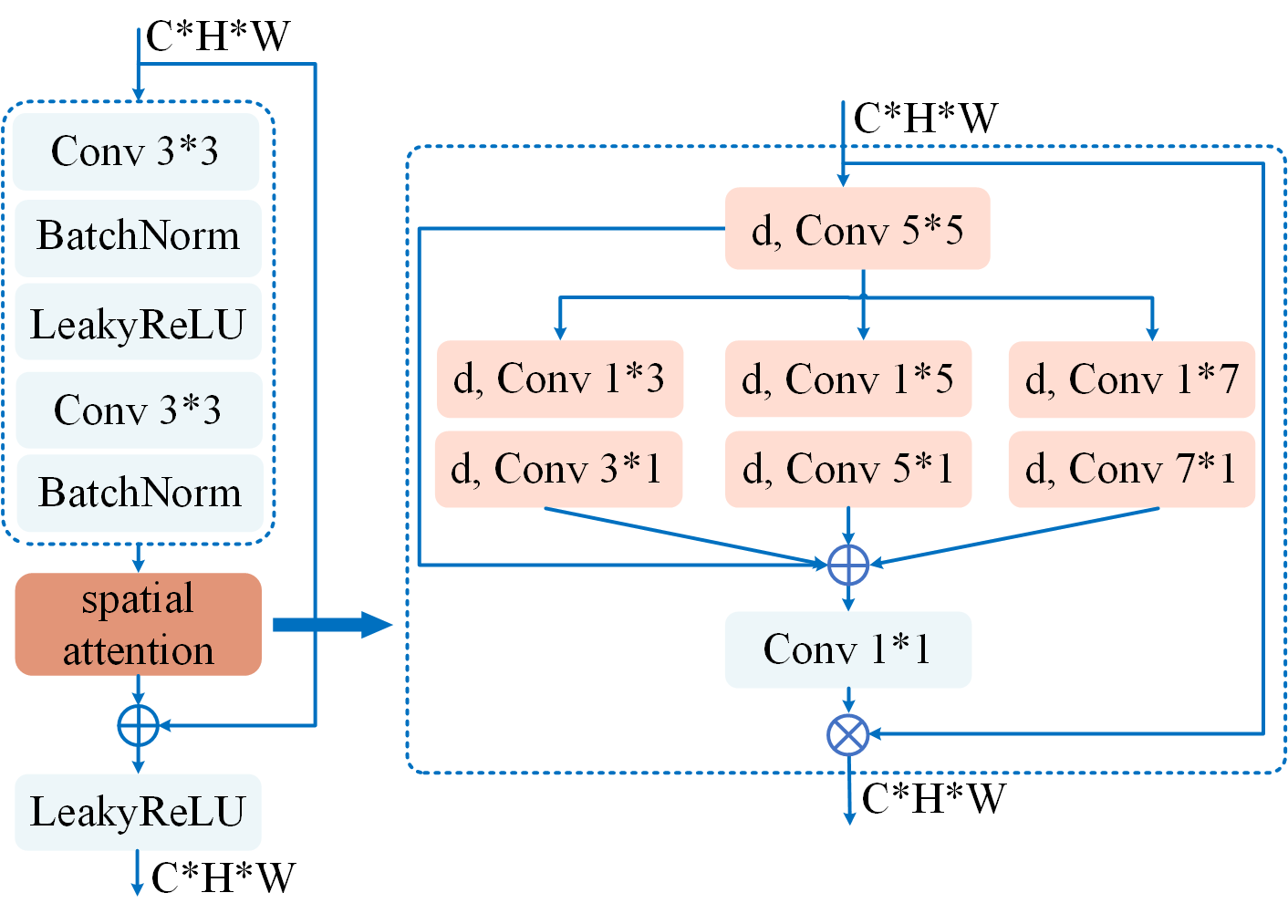}} \\
    \subfloat[The Dual Feature Fusion Block]{
        \label{figure_5b}
        \includegraphics[width = 0.70\linewidth,scale=1.0]{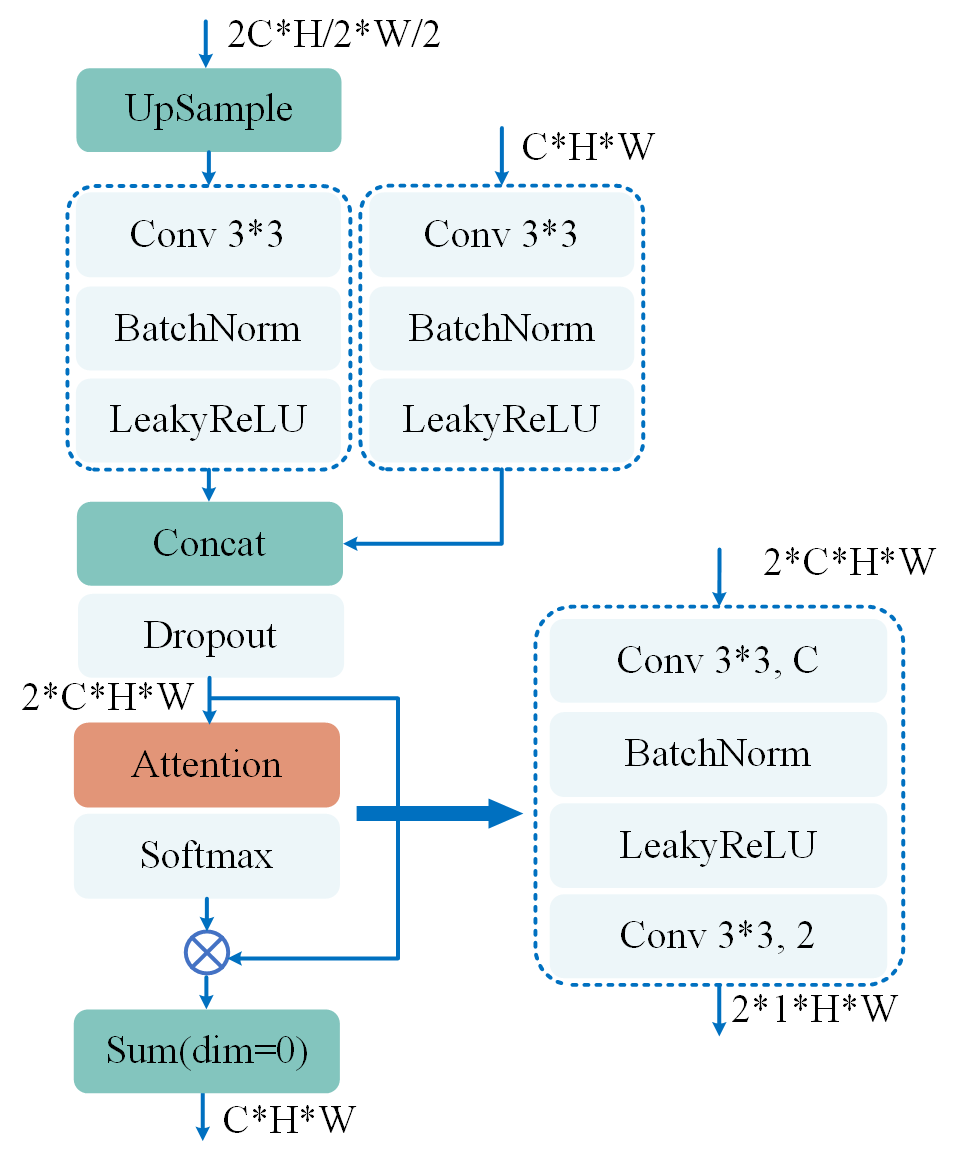}}  
\caption{The Spatial Attention Block (SAB) and the Dual Feature Fusion Block (DFFB).}
\label{fig5}
\end{figure}
\subsection{Network Structure}
\subsubsection{Multi-Branch Networks}
We adopt a hybrid strategy to exploit the complementary strengths of diverse point cloud representations, as shown in Fig.~\ref{fig3}. The network consists of three branches: the point-based branch, the BEV branch, and the RV branch.
For the point-based branch, a three-layer MLP is utilized to extract features from the raw point clouds. 
For the BEV and RV branches, we employ a 2D fully convolutional neural network to extract motion and appearance features, respectively. 
To mitigate boundary artifacts caused by the back-projection process while preserving the detailed information from the raw point clouds, we apply adaptive feature aggregation from all three branches using the PF module.
For the BEV branch, we construct a $(N+1)C \times H_{bev} \times W_{bev}$ tensor by concatenating the BEV image of the current frame with the feature differences from previous scans, where $N$ represents the number of previous scans and $C$ denotes the number of feature channels. In the RV branch, the output of the point-based branch is transformed into an RV image, forming a $C \times H_{rv} \times W_{rv}$ tensor.

\subsubsection{Fully Convolutional Network}
The fully convolutional network is built upon the encoder-decoder framework of ResNet \cite{he2016deep}. 
For the encoder, we employ the Dual Down-sampling Block (DDB) to retain fine-grained information while progressively reducing the feature map's scale. Drawing inspiration from LeNet \cite{ding2023lenet}, we incorporate a Spatial Attention Block (SAB) at the final layer of each stage (see Fig.~\ref{fig5}(a)). This block utilizes convolutional kernels of sizes $3 \times 3$, $5 \times 5$, and $7 \times 7$ to extract features at multiple receptive fields, enabling the capture of both local and global contextual information.
To enhance model efficiency, we decompose the $3 \times 3$ convolutional kernel into $1 \times 3$ and $3 \times 1$ kernels, among others. 

Given the diverse motion modes, velocities, morphologies, and sizes of objects, along with the challenge posed by the uneven distribution of point cloud density---complicating the recognition of distant and small targets---we design a Multi-Kernel Convolutional Neural Network block (MKCNN). Operating with a stride of 2 and utilizing kernels of varying sizes, the MKCNN effectively captures a range of motion patterns and improves segmentation accuracy, especially for distant and small objects. 
The MKCNN block is shown in Fig.~\ref{fig4}. It receives the feature maps produced by the preceding encoder stage and outputs refined multi-scale features for the subsequent fusion stage.
While the MKCNN block draws inspiration from the multi-branch design of the Inception architecture, it fundamentally differs in both implementation and task alignment. 
First, MKCNN is specifically tailored for dynamic point cloud segmentation, where data sparsity and long-range target recognition are critical. To this end, it employs directionally decomposed convolutions (e.g., $1 \times 7$ followed by $7 \times 1$), which significantly expand the receptive field while reducing computational overhead. 
Second, MKCNN integrates a diverse set of kernel sizes ($1 \times 1$, $3 \times 3$, $5 \times 5$, $7 \times 7$), enabling multi-scale motion modeling. Larger kernels help capture broad spatial context in distant and sparse regions, while smaller kernels focus on fine-grained details in near-range and dense areas. 
Therefore, unlike Inception---which primarily serves image classification in dense domains---MKCNN is purpose-built for 3D temporal perception, offering both task-specific efficiency and adaptability to complex dynamic environments. 

For the decoder, we employ Dual Feature Fusion Blocks (DFFB) shown in Fig.~\ref{fig5}(b), which concatenate the feature maps of different layers and adopt an attention mechanism to adaptively extract motion and appearance features. 

\subsubsection{Point Fusion module}
\begin{figure}[!t]
    \centering    
    \includegraphics[width = 0.65\linewidth,scale=0.4]{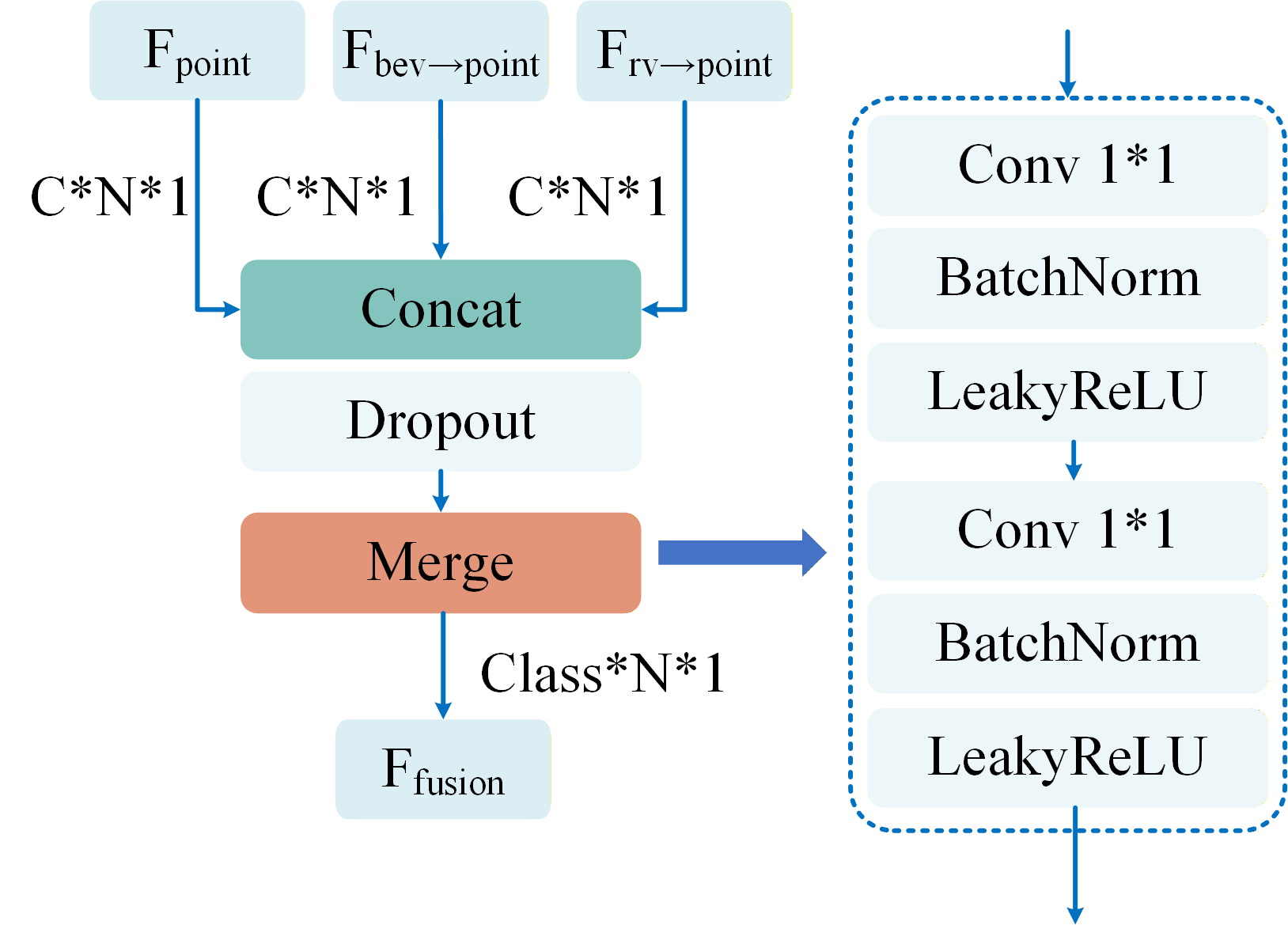}
\caption{The Point Fusion (PF) module.}
\label{fig6}
\end{figure}

\begin{figure}[!t]
    \centering    
    \includegraphics[width = 0.95\linewidth,scale=0.8]{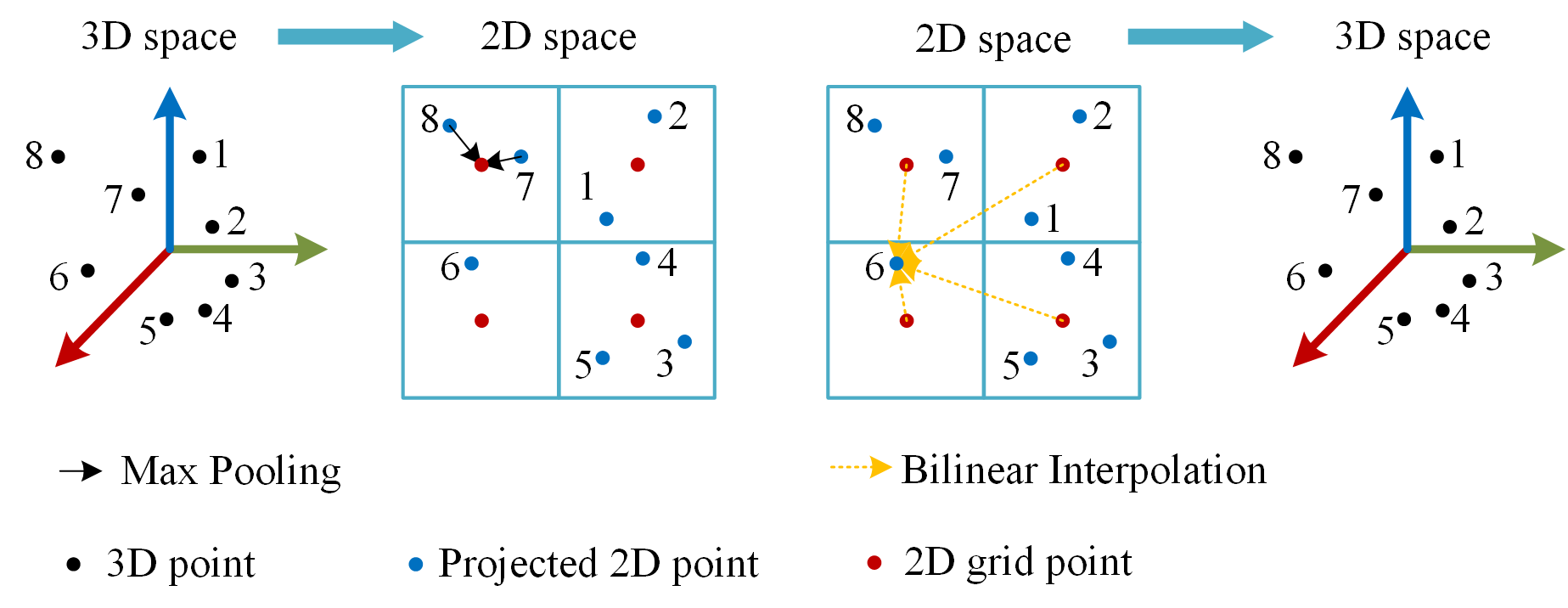}
\caption{The Point to Grid and Grid to Point.}
\label{fig11}
\end{figure}

To achieve fine-grained point-level perception and integrate multi-view features, we design a Point Fusion (PF) module that maps the 2D feature maps from the BEV and RV branches back onto the original 3D points, enabling unified representation across views. As illustrated in Fig.~\ref{fig6}, the PF module takes the point-based features $F_{point}$ and the back-projected BEV and RV features $F_{\text{bev} \rightarrow \text{point}}$ and $F_{\text{rv} \rightarrow \text{point}}$ as inputs and outputs the fused features $F_{fusion}$, which are then fed to the prediction heads. This design fuses point-, BEV-, and RV-based features while preserving geometric details and maintaining the deployment efficiency and training consistency of the end-to-end framework. 
\textbf{Grid to Point.} 
To propagate semantic and motion cues from BEV and RV branches back to each 3D point, we apply bilinear interpolation to map the 2D feature maps into point-level representations, as shown in Fig.~\ref{fig11}. 
Specifically, for each point $p_i$ with projected coordinate $(u_i, v_i)$ on the 2D grid, we retrieve the four nearest neighbors $(u_j, v_j)$ and compute the interpolated feature as:
\begin{equation}
    F_{i}^{interp}=\sum_{j=1}^{4}w_j\cdot F_{grid}(u_j,v_j),
\end{equation}
where $w_j$ denotes the interpolation weights, satisfying $\sum_j w_j = 1$. 
This interpolation process is independently applied to both the BEV and RV branches, resulting in point-level features $F_{\text{bev} \rightarrow \text{point}}$ and $F_{\text{rv} \rightarrow \text{point}}$. These features are then combined with the original point-based features $F_{\text{point}}$ to enable multi-view spatiotemporal feature integration. 
\textbf{Feature Fusion.}
The point-level features from the three branches are concatenated and passed through a lightweight fusion network to generate the final fused representation:
\begin{equation}
    F_{fusion}=Merge(concat(F_{point},F_{\text{bev} \rightarrow \text{point}},F_{\text{rv} \rightarrow \text{point}})),
\end{equation}
Here, $\text{Merge}(\cdot)$ denotes a lightweight fusion module, which can be implemented using a linear layer or an attention-based mechanism to effectively capture complementary information across multiple representations.

\subsection{Loss Function}

Given the complex spatiotemporal structures and rich semantic information present in dynamic environments---and the important role of auxiliary tasks in improving model interpretability and robustness---we introduce an auxiliary semantic branch alongside the main motion segmentation head. 
Taking the SemanticKITTI\cite{behley2019semantickitti} dataset as an example, motion labels consist of two categories: \textit{moving} and \textit{non-moving}, while semantic labels cover 19 fine-grained object classes. 
Incorporating semantic supervision not only provides structural priors that facilitate motion understanding, but also alleviates the challenge of feature misalignment across different spatial scales. 

To jointly optimize both tasks, we formulate the total training loss as a weighted sum of the motion segmentation loss and the semantic supervision loss:  
\begin{equation}
    L_{\text{total}} = \sum_{i \in \{\text{mos}, \text{semantic}\}} \alpha_i * L_i,
\end{equation}
where $\alpha_{mos}$ and $\alpha_{semantic}$ are the weighting factors for the motion and semantic losses, respectively. 
In our experiments, we set both coefficients to $1.0$, a configuration that consistently yields stable convergence and strong performance throughout training. 

For each sub-task loss term $L_i$, we adopt a combination of weighted cross-entropy loss $L_{wce}$ and Lov$\acute{a}$sz-Softmax loss to simultaneously address class imbalance and improve segmentation accuracy around object boundaries: 
\begin{equation}
    L_{\{\text{mos}, \text{semantic}\}} = w_{1}*L_{wce} + w_{2}*L_{ls},
\end{equation}
where $w_{1}$ and $w_{2}$ denote the weights of the weighted cross-entropy and Lov$\acute{a}$sz losses, respectively. 
Following common practices in segmentation tasks, we set $w_1=1.0$ and $w2=1.0$. 
This balanced configuration ensures both class-level discriminability and boundary sensitivity, thereby improving the overall segmentation performance. 

To address the imbalance distribution of classes in the dataset, the cross-entropy loss with frequency weights is defined as follows:
\begin{equation}
    L_{wce}(y,\hat{y} )=-\sum_{i}\frac{1}{\sqrt{f_{i}}}p(y_{i})log(p(\hat{y_{i}})), 
\end{equation}
where $y_{i}$ and $\hat{y_{i}}$ are the true and predicted labels, respectively, and $f_{i}$ is the frequency of the $i$-th class.

To optimize the intersection-over-union (IoU), the Lov$\acute{a} $sz-Softmax loss is employed to maximize the IoU score, which is defined as follows:
\begin{equation}
    \begin{aligned}
    L_{ls} & =\frac{1}{\left | C \right | } \sum_{c\in C}\overline{\Delta J_{c}}(m(c)),  \\
    m_{i}(c) & = \begin{cases}1-x_{i}(c), & \text{ if } c=y_{i} \\x_{i}(c), & \text{ otherwise } \end{cases},
    \end{aligned}
\end{equation}
where $\left | C \right |$ is the number of classes, $\overline{\Delta J_{c}}$ represents the Lov$\acute{a}$sz extension of the Jaccard index,
$x_{i}(c)\in [0,1]$ denotes the predicted probability that pixel $i$ belongs to class $c$, and $y_{i}\in \{1,\dots,\left | C \right |\}$ denotes the ground truth class label of pixel $i$.

\begin{figure*}[!t]
    \centering    
    \includegraphics[width = \linewidth,scale=1.0]{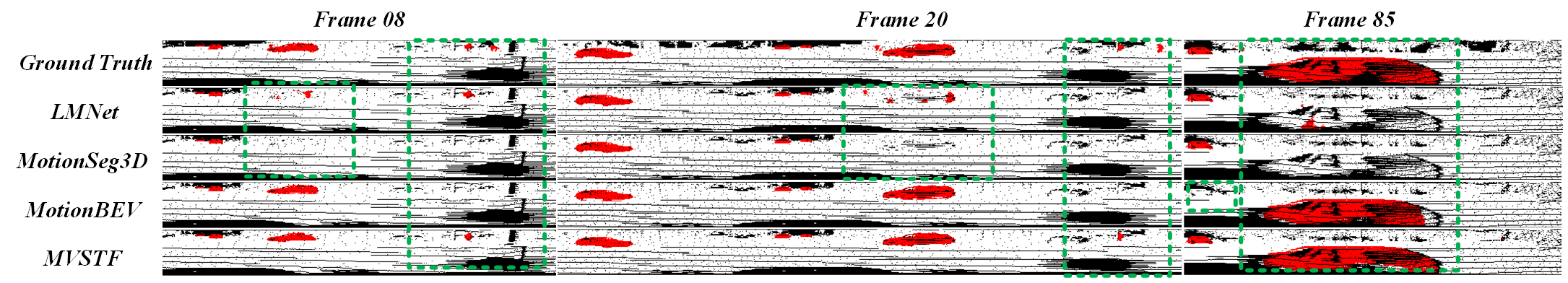}
\caption{Qualitative comparison of the segmentation results of the various methods in the range view. Red represents moving point clouds and gray represents static point clouds. There are three main types of detection errors: incomplete object segmentation, undetected motion objects, and incorrect motion classification.}
\label{fig7}
\end{figure*}

\begin{figure*}[!t]
    \centering    
    \includegraphics[width = \linewidth,scale=1.0]{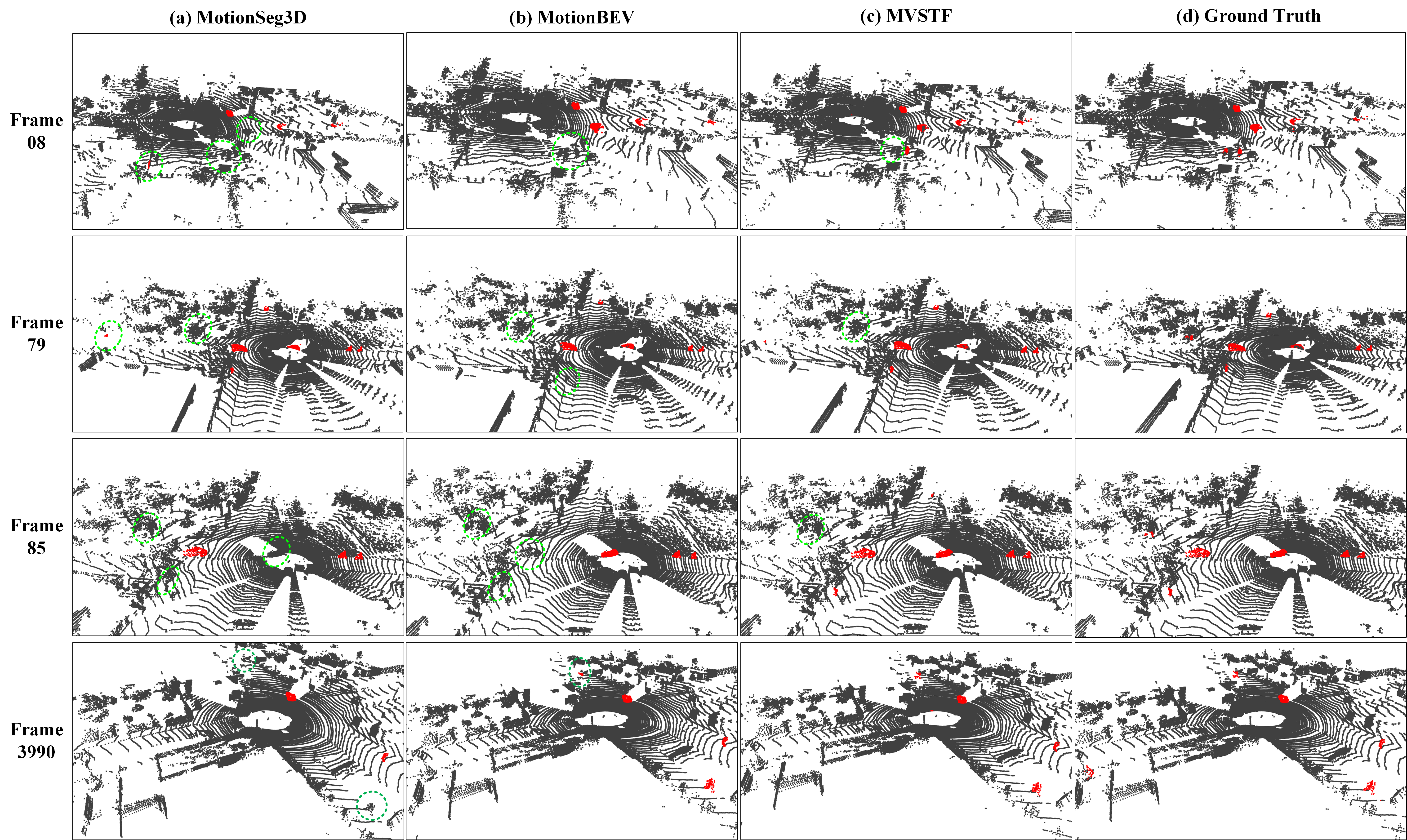}
\caption{Qualitative comparison of MVSTF with MotionSeg3D\cite{sun2022efficient}, and MotionBEV\cite{zhou2023motionbev} on the SemanticKITTI-MOS validation set. Green circles highlight false predictions.}
\label{fig8}
\end{figure*}

\begin{figure*}[!t]
    \centering    
    \includegraphics[width = \linewidth,scale=1.0]{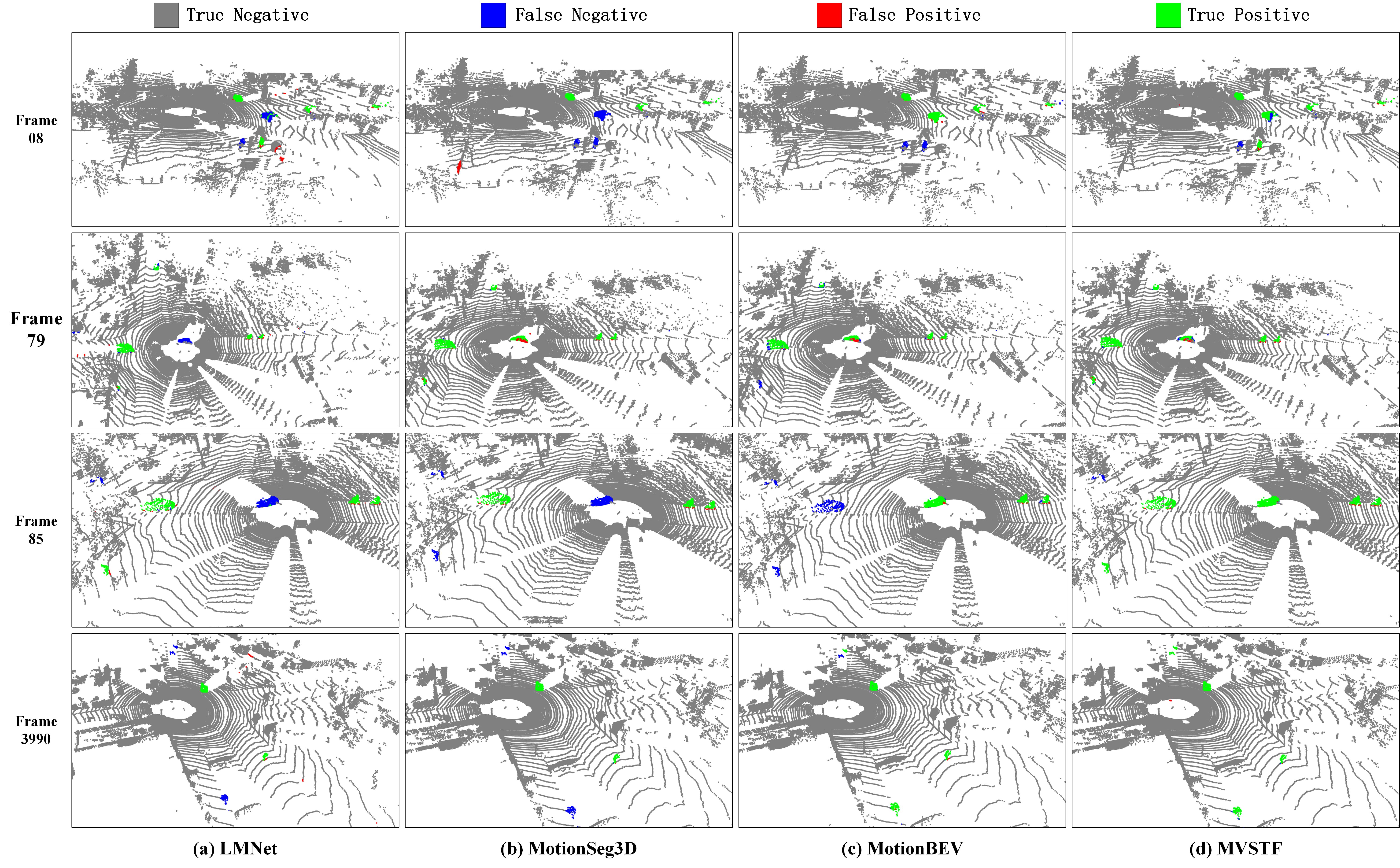}
\caption{Qualitative analysis of different methods for LiDAR-MOS on the SemanticKITTI-MOS validation set.}
\label{fig9}
\end{figure*}

\section{Experiments}
\subsection{Experiment Setups}
\subsubsection{Datasets}
The SemanticKITTI dataset\cite{behley2019semantickitti,chen2021moving} is collected from a city in Germany using high-resolution and high-density Velodyne HDL-64E S3 LiDAR sensors. 
It includes $26$ semantic categories, consisting of $6$ moving classes, $19$ static classes, and $1$ outlier class. 
The dataset comprises $22$ sequences, with sequences $00$-$07$ and $09$-$10$ serving as the training set, sequence 08 as the validation set, and sequences $11$-$21$ used for testing. 
For the MOS task, all point cloud categories are remapped into two classes: moving and non-moving/static. 
To mitigate the issue of imbalanced data distribution, we incorporate an additional KITTI-Road dataset \cite{chen2022automatic}, where sequences $30$-$34$ and $40$ are used for training, and sequences $35$-$39$ and $41$ are reserved for validation. This additional training data improves the model's generalization across different environments. 

To enhance the identification of rare categories, PointPillars \cite{lang2019pointpillars} employs an instance copy-paste strategy. 
Since motion features are derived from LiDAR sequences, the instance copy-paste method needs to be adapted from single-frame to multi-frame contexts. The Sequential Instance Copy-Paste (SICP) strategy \cite{zhang2022sequential} addresses this by considering insertion positions and ray occlusion, ensuring more accurate feature augmentation across frames.

\subsubsection{Evaluation Metrics}
To quantitatively evaluate the MOS performance, we adopt the IoU metric\cite{everingham2010pascal} on the moving object for measurement, which is defined as follows:  
\begin{equation}
    \text{IoU}_{\text{MOS}}=\frac{\text{TP}}{\text{TP}+\text{FP}+\text{FN}},
\end{equation}
where $TP$, $FP$, and $FN$ denote true positive, false positive, and false negative predictions for the moving class, respectively.
In addition, we report the Mean IoU in Table~\ref{tab:static_moving_metrics}, which is computed as the average of the per-class IoUs over the moving and non-moving classes:
\begin{equation}
    \text{Mean IoU} = \frac{\text{IoU}_{\text{moving}} + \text{IoU}_{\text{static}}}{2},
\end{equation}
where $\text{IoU}_{\text{moving}}$ corresponds to the $\text{IoU}_{\text{MOS}}$ defined above and $\text{IoU}_{\text{static}}$ is computed analogously for the static class.

\subsection{Comparisons with State-of-the-Art methods}
We evaluate and compare the proposed method on the SemanticKITTI-MOS benchmark with other methods, including 
LiMoSeg\cite{mohapatra2021limoseg}, LMNet\cite{chen2021moving}, MotionSeg3D\cite{sun2022efficient}, RVMOS\cite{kim2022rvmos}, 4DMOS\cite{mersch2022receding}, InsMOS\cite{wang2023insmos}, MotionBEV\cite{zhou2023motionbev}, StreamMOS\cite{li2024streammos}, MF-MOS\cite{cheng2024mf}, LiDAR-IMU-GNSS\cite{li2023multi}, MambaMOS\cite{zeng2024mambamos}. The quantitative evaluation results of the different methods on the SemanticKITTI dataset are shown in Table~\ref{tab1} and Table~\ref{tab:static_moving_metrics}. 
Table~\ref{tab1} reports the quantitative results on the SemanticKITTI-MOS benchmark. MVSTF achieves $84.5\%$ and $77.6\%$ IoU on the validation and hidden test sets, respectively, the best results among all compared methods. It surpasses MambaMOS\cite{zeng2024mambamos} by $2.2$ and $2.0$ percentage points, and LiDAR-IMU-GNSS\cite{li2023multi} by $5.5$ and $2.7$ percentage points, on the validation and test sets, respectively. Additional precision, recall, and mean IoU results are provided in Table~\ref{tab:static_moving_metrics}. The comparison under the combined SemanticKITTI-MOS and KITTI-Road training protocol is presented in Section IV-E.

\begin{table}[htbp]
\centering
\caption{Comparison With State-of-the-Art Methods on the SemanticKITTI-MOS Benchmark.}
\begin{tabular}{c|cc}
\toprule
\multirow{2}{*}{\textbf{Method}} & \multicolumn{2}{c}{$\mathbf{IoU_{MOS}(\%)}$} \\
\cmidrule(lr){2-3}
& \textbf{Validation 08} & \textbf{Test 11-21} \\
\midrule
LiMoSeg \cite{mohapatra2021limoseg} & 52.6 & - \\
LMNet \cite{chen2021moving} & 67.1 & 54.5 \\
MotionSeg3D \cite{sun2022efficient} & 71.4 & 64.9 \\
4DMOS \cite{mersch2022receding} & 77.2 & 65.2 \\
MotionBEV \cite{zhou2023motionbev} & 76.5 & 69.7 \\
LiDAR-IMU-GNSS \cite{li2023multi} & 79.0 & 74.9 \\
StreamMOS\cite{li2024streammos} & 78.3 & 73.1 \\
MambaMOS \cite{zeng2024mambamos} & 82.3 & 75.6 \\
MVSTF & \textbf{84.5} & \textbf{77.6} \\
\midrule
LMNet* \cite{chen2021moving} & 63.8 & 60.5   \\
MotionSeg3D* \cite{sun2022efficient} & 69.3 & 70.2 \\
InsMOS* \cite{wang2023insmos} & 69.4 & 75.6   \\
MF-MOS*\cite{cheng2024mf}  & - & \textbf{76.7} \\
MVSTF* & \textbf{85.60} & 76.1 \\
\bottomrule
\end{tabular}
\tablenote{* indicates methods trained on both the SemanticKITTI and KITTI-Road datasets.}
\label{tab1}
\end{table}

\begin{table}[htbp]
\centering
\caption{Performance Comparison of Different Methods on Moving Object Segmentation. All Values Are Reported in Percentages.}
\label{tab:mos_metrics_only}
\renewcommand{\arraystretch}{1.2}
\setlength{\tabcolsep}{4.5pt}
\begin{tabular}{l|ccc|c}
\toprule
\textbf{Method} & $\mathbf{IoU_{MOS}}$ & $\mathbf{Precision}$ & $\mathbf{Recall}$ & \textbf{Mean IoU} \\
\midrule
LMNet \cite{chen2021moving}        & 67.11 & 87.58 & 74.17 & 83.49 \\
MotionSeg3D \cite{sun2022efficient}  & 69.76 & 88.15 & 76.98 & 84.82 \\
InsMOS \cite{wang2023insmos}       & 70.72 & 92.95 & 74.72 & 85.30 \\
MotionBEV \cite{zhou2023motionbev}    & 77.88 & 93.52 & 82.33 & 88.90 \\
MVSTF            & 84.51 & 96.52 & 87.16 & 92.23 \\
MVSTF\textsuperscript{*} & \textbf{85.60} & \textbf{96.21} & \textbf{88.59} & \textbf{92.77} \\
\bottomrule
\end{tabular}
\tablenote{All methods are reproduced locally using their official code and publicly available configurations. * indicates methods trained on both the SemanticKITTI and KITTI-Road datasets.}
\label{tab:static_moving_metrics}
\end{table}

The segmentation results of various methods in RV and BEV are shown in Fig.~\ref{fig7} and Fig.~\ref{fig8}. It can be observed that LMNet\cite{chen2021moving}, MotionSeg3D\cite{sun2022efficient}, and MotionBEV\cite{zhou2023motionbev} have issues of incomplete segmentation of dynamic objects, undetected or incorrectly detected dynamic objects, and inferior recognition ability for distant and small objects. 
To distinguish the segmentation effectiveness of different algorithms more explicitly, \revred{the confusion matrix comparing the point cloud predictions with the ground truth is shown in Fig.~\ref{fig9}}.
It is observed that some methods suffer from boundary ambiguity, incorrect classification of remote small objects, and absence of object detection, whereas MVSTF effectively solves such problems and significantly boosts the performance. 
Since the density distribution of the LiDAR point cloud is not uniform, the farther the object is from the LiDAR, the sparser the point cloud becomes. Table~\ref{tab2} analyzes in detail the performance metrics for close, medium, and far range. 
\revred{Most methods perform better at close range than at medium and far ranges.} 
At close range, MVSTF achieves the best performance relative to other methods. 
At medium range, MVSTF and MambaMOS\cite{zeng2024mambamos} have comparable performance. 
\begin{table*}[htbp]
\centering
\caption{MOS Performance on the SemanticKITTI-MOS Validation Set at Different Ranges. R Denotes Recall and P Denotes Precision.}
\label{tab2}
\begin{tabular}{c|ccc|ccc|ccc}
\toprule
\multirow{2}{*}{\textbf{Method}} & \multicolumn{3}{c|}{\textbf{Close ($<$20m)}} & \multicolumn{3}{c|}{\textbf{Medium ($\geq$20m, $<$ 50m)}} & \multicolumn{3}{c}{\textbf{Far ($\geq$ 50m)}} \\
\cmidrule(lr){2-4} \cmidrule(lr){5-7} \cmidrule(l){8-10}
& $\mathbf{IoU_{MOS}}$ & $\mathbf{R}$ & $\mathbf{P}$ & $\mathbf{IoU_{MOS}}$ & $\mathbf{R}$ & $\mathbf{P}$ & $\mathbf{IoU_{MOS}}$ & $\mathbf{R}$ & $\mathbf{P}$ \\
\midrule
LMNet \cite{chen2021moving} & 70.72 & 76.89 & 89.80 & 43.88 & 54.30 & 69.56 & 0.00 & 0.00 & - \\
MotionSeg3D \cite{sun2022efficient} & 72.26 & 79.40 & 88.94 & 52.11 & 59.32 & 81.08 & 5.30 & 5.30 & \textbf{100.00}  \\
InsMOS \cite{wang2023insmos} & 75.29 & \underline{88.78} & 83.21 & 57.67 & 66.81 & 80.84 & 10.88 & 10.89 &  \underline{98.63} \\
MotionBEV \cite{zhou2023motionbev} & 80.85 & 85.40 & 93.81 & 56.35 & 59.90 & 90.50 & 0.00 & 0.00 & - \\
MF-MOS \cite{cheng2024mf}  & 79.31 & 84.98 & 92.23 & 54.67 & 64.10 & 78.81 & 47.97 & 50.08 & 91.94 \\
MambaMOS \cite{zeng2024mambamos} & 83.69 & 87.30 & 95.29 & \textbf{72.48} &  78.24 &  \underline{90.78} & \textbf{94.44} & \textbf{97.73} & 96.56 \\
MVSTF & \underline{86.33} & 88.33 & \textbf{97.45} & 71.90 & \underline{78.55} & 89.47 & \underline{82.04} & \underline{86.38} & 94.22 \\
MVSTF* & \textbf{87.55} & \textbf{89.75} & \underline{97.27} & \underline{72.28} & \textbf{80.02} & 88.19 & 77.81 & 80.64 & 95.69 \\
\bottomrule
\end{tabular}
\end{table*}

In long-range scenarios (beyond $50m$), single-view methods such as LMNet \cite{chen2021moving}, MotionSeg3D \cite{sun2022efficient}, and MotionBEV \cite{zhou2023motionbev} exhibit a significant drop in performance, often failing to accurately segment moving objects. 
This degradation stems from two key factors. 
First, long-range LiDAR data are inherently sparse and more susceptible to noise and occlusion, severely limiting the representational capacity of single-perspective features. 
Second, these methods typically rely on simplified temporal cues---such as RV depth residuals or polar-coordinate BEV height differences---which are inadequate for capturing subtle motion dynamics in distant regions. 
In contrast, MVSTF maintains superior performance at long range, benefiting from complementary information captured through both BEV and RV projections. 
Specifically, MVSTF leverages inter-frame BEV feature differencing to explicitly model temporal motion cues. 
The spherical projection component is designed to help mitigate occlusion and angular distortion inherent in single-view representations. 
Furthermore, the multi-view fusion strategy aims to complement point-level features, thereby improving semantic consistency and boundary delineation across regions with varying spatial densities. 
This comprehensive fusion across spatial, temporal, and feature dimensions enables MVSTF to achieve reliable motion segmentation even in sparse and occluded regions. 

\begin{table*}[htbp]
\centering
\caption{Ablation Study on Each Component in MVSTF on the SemanticKITTI-MOS Validation Set. 
DDB: Dual Down-sampling Blocks, SAB: Spatial Attention Blocks, MKCNN: Multi-Kernel Convolutional Neural Network Blocks, DFFB: Dual Feature Fusion Blocks. 
}
\begin{tabular}{@{}c|cc|cccc|cc|c@{}}
\toprule
& \multicolumn{2}{c|}{\textbf{Temporal Branch}} & \multicolumn{4}{c|}{\textbf{Component}} & \multicolumn{2}{c|}{\textbf{Head Structure}} \\
\cmidrule(lr){2-3} \cmidrule(lr){4-7} \cmidrule(lr){8-9}
& \textbf{BEV} & \textbf{RV} & \textbf{DDB} & \textbf{SAB} & \textbf{MKCNN} & \textbf{DFFB} & \textbf{One head} & \textbf{Two Head} & $\mathbf{IoU_{MOS}(\%)}$ \\
\midrule
(a) & 2 & - & $\checkmark$ & & & & $\checkmark$ & & 70.27 \\
(b) & - & 2 & $\checkmark$ & & & & $\checkmark$ & & 65.73 \\
(c) & 2 & 1 & $\checkmark$ & & & $\checkmark$ & $\checkmark$ & & 74.83 \\
(d) & 8 & 1 & $\checkmark$ & & & $\checkmark$ & $\checkmark$ & & 78.98 \\
(f) & 8 & 1 & $\checkmark$ & $\checkmark$ & & $\checkmark$ & $\checkmark$ & & 81.09 \\
(g) & 8 & 1 & $\checkmark$ & $\checkmark$ & & $\checkmark$ & & $\checkmark$ & 82.43 \\
(h) & 8 & 1 & $\checkmark$ & $\checkmark$ & $\checkmark$ & $\checkmark$ & $\checkmark$ & & 83.67 \\
(i) & 8 & 1 & $\checkmark$ & $\checkmark$ & $\checkmark$ & $\checkmark$ & & $\checkmark$ & \textbf{84.51} \\
(j) & 8 & 1 & $\checkmark$ & $\checkmark$ & $\checkmark$ & $\checkmark$ & & $\checkmark$ & \textbf{85.60*} \\
\bottomrule
\end{tabular}
\label{tab3}
\tablenote{* indicates methods trained on both the SemanticKITTI and KITTI-Road datasets. “One Head” denotes motion prediction only, and “Two Head” denotes joint motion and semantic prediction.}
\end{table*}

\begin{table}[htbp]
\caption{Ablation Study About Different Input Types of the BEV Branch on the SemanticKITTI-MOS Validation Set.}
    \begin{center}
    \setlength{\tabcolsep}{6.5mm}{
    \begin{tabular}{l|l|l}
    \toprule[1pt]
      & \textbf{Input (N=8)}  & $\mathbf{IoU_{MOS} (\%)}$\\ 
    \midrule[0.5pt]
    (a) & Continuous frames     & 76.37 \\    
    (b) & Height Residuals      & 79.09 \\ 
    (c) & Feature Differences    & \textbf{84.51}  \\
    \bottomrule[1pt] 
    \end{tabular}
    }
\label{tab4}
\end{center}
\end{table}

\begin{table}[htbp]
\centering
\caption{Ablation Study About the Number of Historical Frames on the SemanticKITTI-MOS Validation Set.}
\label{tab5}
\setlength{\tabcolsep}{4pt}
\begin{tabular}{@{}c|cc|cc@{}}
\toprule
\multirow{2}{*}{\textbf{Frame $N$}} & \multicolumn{2}{c|}{\textbf{Height Residuals}} & \multicolumn{2}{c}{\textbf{Feature Differences}} \\
\cmidrule(lr){2-3} \cmidrule(l){4-5}
                    & $\mathbf{IoU_{MOS} (\%)}$ & \textbf{runtime(ms)} & $\mathbf{IoU_{MOS} (\%)}$ & \textbf{runtime(ms)} \\ \midrule
2                 & 72.71        & 68.22   & 81.60        & 190.94  \\
4                 & 74.92        & 76.22   & 82.76        & 223.36  \\
6                 & 75.60        & 81.85   & 82.96        & 233.66  \\
8                 & \textbf{79.09} & \textbf{94.99} & \textbf{85.60*} & \textbf{238.92} \\
10                & 74.65        & 102.92  & 81.32        & 269.12  \\
12                & 75.94        & 135.11  & 79.36        & 289.99 \\ \bottomrule
\end{tabular}
\tablenote{* indicates methods trained on both the SemanticKITTI and KITTI-Road datasets.}
\end{table}

\begin{table}[htbp]
\centering
\caption{Comparison with state-of-the-art methods on the KITTI-Road dataset.}
\begin{tabular}{lc} 
\toprule
\textbf{Method} & $\mathbf{IoU_{MOS} (\%)}$ \\
\midrule
LMNet \cite{chen2021moving}  & 87.4   \\
MotionBEV \cite{zhou2023motionbev}  & 80.5  \\
InsMOS \cite{wang2023insmos}  & 83.9  \\
MF-MOS \cite{cheng2024mf}  & 87.9 \\
MambaMOS \cite{zeng2024mambamos} &  89.4 \\
\midrule
MVSTF  & \textbf{89.8} \\
\bottomrule
\end{tabular}
\label{tab6}
\end{table}

\subsection{Ablation Study}
We perform ablation studies on the SemanticKITTI-MOS validation set to verify the effectiveness of the different components of the proposed method. As shown in Table~\ref{tab3}, it can be observed that: 
(1) The BEV branch is more suitable than the RV branch to extract the motion properties of point clouds (a,b);   
(2) The multi-view is better than the single view (b,c), which effectively solves the occlusion and boundary blur problems; 
(3) The appropriate increase in the number of BEV feature differences frames N improves the MOS performance (c,d); 
(4) The introduction of SAB significantly improves the MOS performance and realizes the spatial perception of different feature pyramid layers (d,f); 
(5) The auxiliary semantic head accelerates parameter convergence and enhances the model's interpretability and robustness (f,g and h,i); 
(6) The MKCNN module adopts different sizes of convolutional kernels to capture various sizes of moving objects (f,h and g,i). 

The motion attributes of point clouds are extracted from successive frames with temporal information, and RV images are sensitive to distance variations, so the depth residuals provide relatively limited gains in performance. 
\revred{Since most moving objects in autonomous-driving scenarios primarily move in the $x$-$y$ plane}, the feature differences of the BEV images effectively highlight motion-related changes while suppressing static content, which improves the extraction of motion features.
Moreover, the projection distortion of a single view may confuse adjacent, similar, or occluded points of view, whereas multi-view fusion enables complementary information. 
The attention mechanism in SAB focuses on the crucial features, and applying SAB to different feature layers can enrich the receptive field of the network. 
The auxiliary semantic head utilizes the loss function of the semantic segmentation task to improve the network performance. 
The MKCNN module can effectively recognize objects of various shapes and sizes, especially distant and small objects. 

Furthermore, we compare the effects of taking three different types of inputs in BEV on the MOS performance, as shown in Table~\ref{tab4}. 
It can be seen that computing feature differences in BEV space performs better compared to directly connecting consecutive frames or computing height residuals among multiple frames.

\subsection{History Frames and Runtimes}
To select a suitable number of historical frames, we investigate the impact of the previous frame number $N$ on MOS performance and runtime, as shown in Table~\ref{tab5}.  All experiments are conducted on the SemanticKITTI validation set using a single NVIDIA RTX 3090 GPU. 
We observe that an appropriate increase in the number of historical frames $N$ can improve the model's ability to capture motion cues in point clouds.
Nevertheless, if the number of historical frames is too high, the object's motion state may shift significantly during this window, leading to degraded segmentation accuracy. 

The LiDAR data in SemanticKITTI is collected by a Velodyne HDL-64E S2 sensor operating at $10 Hz$, implying a $100 ms$ interval between consecutive scans. 
In this context, we propose two temporal modeling strategies to meet varying application requirements: 
\begin{itemize}
\item \textbf{Feature Difference Strategy} (with $N=8$) achieves the highest $IoU_{MOS}$ of 85.60\%, but incurs an average inference time of $238.92ms$, exceeding the scan interval. This makes it suitable for high-precision offline tasks such as ghost artifact removal in HD map construction. 
\item \textbf{Height Residual Strategy} reduces the computational load significantly, achieving 79.09\% $IoU_{MOS}$ with an inference time of only $94.99 ms$---well within the sensor's scan cycle---thus enabling real-time deployment on resource-constrained platforms like in-vehicle or edge systems. 
\end{itemize}
Based on these observations, we introduce a dual-mode temporal modeling framework: 
The feature difference module targets accuracy-critical offline scenarios. 
The height residual module prioritizes real-time efficiency for online perception. 
This dual-path design offers a flexible trade-off between accuracy and efficiency, enabling adaptive deployment across diverse environments. 

\subsection{Generalization}
To examine the generalization of MVSTF, we train the model jointly on the SemanticKITTI-MOS\cite{chen2021moving} and KITTI-Road\cite{chen2022automatic} training sets. 
As shown in rows (i) and (j) of Table~\ref{tab3}, the additional KITTI-Road training data improves IoU$_{MOS}$ from $84.51\%$ to $85.60\%$ on the SemanticKITTI validation set, an improvement of $1.09$ percentage points. Combined with Table~\ref{tab2}, MVSTF achieves its largest gains in the close and medium ranges. 
In Table~\ref{tab1}, MVSTF* outperforms MambaMOS\cite{zeng2024mambamos} by 3.3 percentage points on the SemanticKITTI-MOS validation set\cite{chen2021moving}. On the test set, MVSTF* achieves a score of 76.1\%, exceeding MambaMOS\cite{zeng2024mambamos} by 0.5 percentage points. Furthermore, Table~\ref{tab6} shows that MVSTF* outperforms MambaMOS\cite{zeng2024mambamos} by 0.4 percentage points on the KITTI-Road validation set\cite{chen2022automatic}.

\subsection{Applications}
To assess the practical benefits of MVSTF for localization and mapping, we integrate its motion masks into SuMa++\cite{chen2019suma++} and evaluate the resulting pipeline on KITTI odometry training sequences 00--10. An offline two-pass pipeline is adopted, with all methods sharing the baseline-estimated poses for ego-motion compensation, the backend configuration, and the semantic predictions.

(1) Odometry/SLAM: 
As reported in Table~\ref{tab:odometry_en}, integrating MVSTF reduces the translational error without loop closure from 1.3709\% to 0.8848\%, a relative reduction of 35.46\%, while reducing the rotational error from 0.3411°/100 m to 0.3272°/100 m. With loop closure, the translational error decreases from 1.3475\% to 0.9841\%, corresponding to a relative reduction of 26.97\%. These results demonstrate that MVSTF-based motion filtering further improves translational accuracy beyond the existing semantic constraints.

\begin{figure*}[t]
    \centering
    \includegraphics[width=\textwidth]{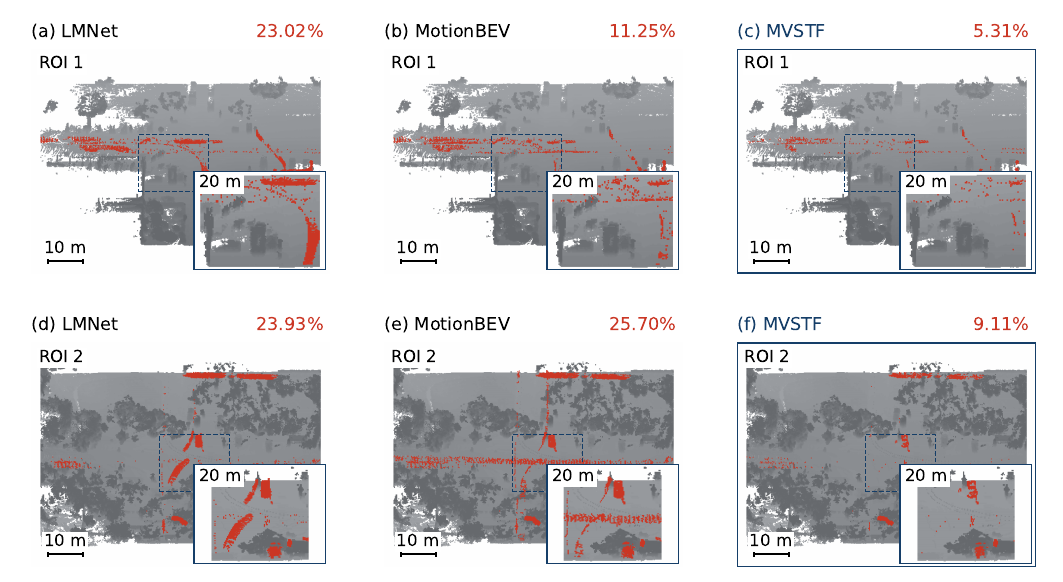}
    \caption{
    3D mapping results for two local regions of sequence 08.
    Panels (a)--(c) show ROI 1, and panels (d)--(f) show ROI 2.
    The columns correspond to LMNet, MotionBEV, and MVSTF,
    respectively.
    Red denotes dynamic-only reference voxels that remain occupied,
    while gray denotes other occupied voxels.
    Blue dashed boxes indicate the regions enlarged in the
    lower-right insets.
    The percentage in the upper-right corner of each panel
    denotes the dynamic-voxel residual rate over the entire
    corresponding ROI; lower values are better.
    All methods use the same baseline-estimated poses,
    voxel grid, and visualization settings.
    }
    \label{fig:mapping_en}
\end{figure*}

(2) 3D Mapping: 
All methods accumulate observations within a range of 2--75 m into 0.2 m voxel maps using the same baseline-estimated poses. As shown in Table~\ref{tab:mapping_en}, MVSTF achieves a dynamic-point removal rate of 95.17\% and a static-point retention rate of 99.9941\%, both the highest among the evaluated learning-based methods. Its dynamic-voxel residual rate is only 4.02\%, representing relative reductions of 57.72\% and 68.14\% compared with LMNet and MotionBEV, respectively, while retaining 99.9743\% of the static-only reference voxels. 
Fig.~\ref{fig:mapping_en} further illustrates these benefits. MVSTF achieves dynamic-voxel residual rates of 5.31\% and 9.11\% in the two local regions, respectively, outperforming both comparison methods. The enlarged views show reduced motion-trail artifacts in road regions while preserving the surrounding static structures. Together, the quantitative and visual results demonstrate a favorable balance between dynamic-observation removal and static-structure preservation.

\begin{table*}[t]
\centering
\caption{Trajectory errors before and after integrating MVSTF into SuMa++.}
\label{tab:odometry_en}
\small
\setlength{\tabcolsep}{8pt}
\renewcommand{\arraystretch}{1.15}
\begin{tabular}{lcccc}
\toprule
\multirow{2}{*}{Configuration}
& \multicolumn{2}{c}{Odometry without loop closure}
& \multicolumn{2}{c}{SLAM with loop closure} \\
\cmidrule(lr){2-3}
\cmidrule(lr){4-5}
& $t_{\mathrm{rel}}$ (\%) $\downarrow$
& $r_{\mathrm{rel}}$ ($^\circ$/100 m) $\downarrow$
& $t_{\mathrm{rel}}$ (\%) $\downarrow$
& $r_{\mathrm{rel}}$ ($^\circ$/100 m) $\downarrow$ \\
\midrule
SuMa++
& $1.3709 \pm 0.0000$
& $0.3411 \pm 0.0000$
& $1.3475 \pm 0.0139$
& $0.3148 \pm 0.0039$ \\
\textbf{SuMa++ + MVSTF}
& $\mathbf{0.8848} \pm 0.0000$
& $0.3272 \pm 0.0000$
& $\mathbf{0.9841} \pm 0.1913$
& $0.3496 \pm 0.0758$ \\
\bottomrule
\end{tabular}

\smallskip
\begin{minipage}{0.98\textwidth}
\footnotesize
\tablenote{
Results are the mean ± sample standard deviation of five backend runs with fixed motion masks on KITTI odometry training sequences 00--10, pooling all valid 100--800 m trajectory segments within each run. The configurations without and with loop closure use online and final optimized trajectories, respectively.
}
\end{minipage}
\end{table*}

\begin{table*}[t]
\centering
\caption{Dynamic filtering and static-structure preservation
of different motion segmentation methods (\%).}
\label{tab:mapping_en}
\small
\setlength{\tabcolsep}{10pt}
\renewcommand{\arraystretch}{1.15}
\begin{tabular}{lcccc}
\toprule
Method
& \makecell{Dynamic-point\\removal $\uparrow$}
& \makecell{Static-point\\retention $\uparrow$}
& \makecell{Dynamic-voxel\\residual $\downarrow$}
& \makecell{Static-voxel\\retention $\uparrow$} \\
\midrule
LMNet
& 92.45
& 99.9788
& 9.51
& 99.9674 \\
MotionBEV
& 85.04
& 99.9914
& 12.61
& \textbf{99.9835} \\
\textbf{MVSTF}
& \textbf{95.17}
& \textbf{99.9941}
& \textbf{4.02}
& 99.9743 \\
\bottomrule
\end{tabular}

\smallskip
\begin{minipage}{0.98\textwidth}
\footnotesize
\tablenote{
Boldface indicates the best result in each column. Static-voxel retention and dynamic-voxel residual are evaluated on static-only and dynamic-only reference voxels, respectively, excluding mixed voxels. All ratios are calculated from the corresponding counts pooled across sequences 00--10 before surfel fusion.
}
\end{minipage}
\end{table*}

\section{Conclusion}
To address the challenges of boundary-blurring, artifacts, single-view limitations, and training complexity commonly encountered in existing LiDAR-based MOS methods, we propose a novel and effective end-to-end high-precision segmentation framework, \textbf{\textit{MVSTF}}. 
First, the sequential point cloud data is preprocessed to generate RV images, BEV images, and BEV feature differences. A three-branch framework is then employed to effectively integrate spatio-temporal features and channel information from multiple views. 
The projection branch utilizes an encoder-decoder architecture, incorporating Dual Down-sampling Blocks (DDB), Spatial Attention Blocks (SAB), Multi-Kernel Convolutional Neural Network blocks (MKCNN), and Dual Feature Fusion Blocks (DFFB). 
An end-to-end framework is established using the point-based branch and the point fusion (PF) module. 
The auxiliary semantic head is introduced to improve the model's interpretability and robustness. 
Comparative experiments and ablation studies on the SemanticKITTI and KITTI-Road datasets validate the effectiveness of the proposed method, which achieves state-of-the-art results on the SemanticKITTI-MOS validation set and the KITTI-Road validation set while remaining competitive on the hidden SemanticKITTI-MOS test set. 

Future work will focus on two main directions. First, we aim to extend MVSTF's multi-sensor fusion capability by incorporating semantic and texture information from cameras, enhancing its perception of fine-grained semantics and motion boundaries. Second, we seek to improve system robustness and adaptability in real-world scenarios, including sensor degradation, adverse weather, and dynamic environments. Additionally, we will explore model compression and lightweight design to enable efficient deployment on edge platforms. 

\bibliography{IEEEabrv,mybib}{}
\bibliographystyle{IEEEtran}
\end{document}